\documentclass{article}

\usepackage{spconf,amsmath,epsfig}
\usepackage{graphicx}
\usepackage{multicol}
\usepackage[export]{adjustbox}
\usepackage{multirow}
\usepackage[table,xcdraw]{xcolor}
\let\OLDthebibliography\thebibliography
\renewcommand\thebibliography[1]{
  \OLDthebibliography{#1}
  \setlength{\parskip}{0pt}
  \setlength{\itemsep}{0pt plus 0.3ex}
}

\usepackage[numbers]{natbib}

\begin{document}\sloppy

\def\x{{\mathbf x}}
\def\L{{\cal L}}

\title{WavEnhancer: Unifying Wavelet and Transformer for Image Enhancement}
%
\twoauthors{Zinuo Li \qquad Xuhang Chen \qquad Chi-Man Pun\sthanks{} }
                           {University of Macau \\
                            Macau}
                           {Shuqiang Wang\sthanks{} }
                           {Shenzhen Institute of Advanced Technology \\
                            Shenzhen}

\maketitle

\begin{abstract}
Image enhancement is a technique that frequently utilized in digital image processing. In recent years, the popularity of learning-based techniques for enhancing the aesthetic performance of photographs has increased. However, the majority of current works do not optimize an image from different frequency domains and typically focus on either pixel-level or global-level enhancements. In this paper, we propose a transformer-based model in the wavelet domain to refine different frequency bands of an image. Our method focuses both on local details and high-level features for enhancement, which can generate superior results. On the basis of comprehensive benchmark evaluations, our method outperforms the state-of-the-art methods.

\end{abstract}

\section{Introduction}

In recent years, digital photography has evolved significantly due to the ongoing improvement of camera sensors. However, image enhancement is still a major issue in post-processing. 

Professional software such as Adobe Photoshop offers interactive and semi-automated capabilities for performing a number of modifications. However, manual modification is challenging for amateurs who lack the processing abilities, time, and aesthetic sensibility to successfully retouch photographs.

\begin{figure}[ht]
    \begin{minipage}[b]{1.0\linewidth}
        \begin{minipage}[b]{.32\linewidth}
            \centering
            \centerline{\includegraphics[height=3cm, frame]{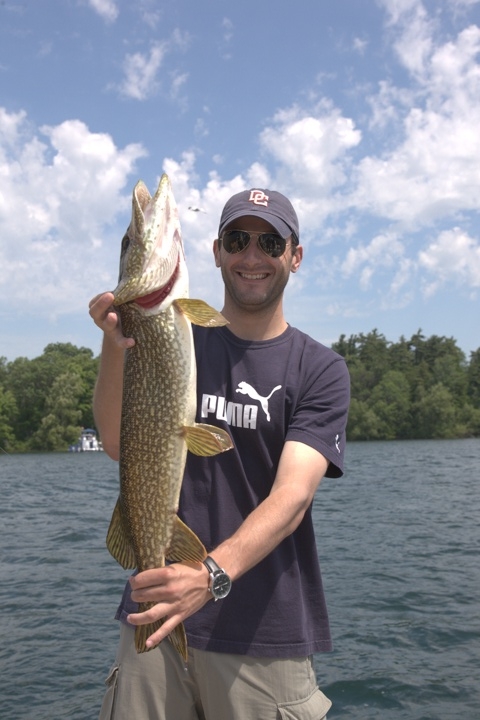}}
            \centerline{(a) Input}\medskip
        \end{minipage}
        \hfill
        \begin{minipage}[b]{0.32\linewidth}
            \centering
            \centerline{\includegraphics[height=3cm, frame]{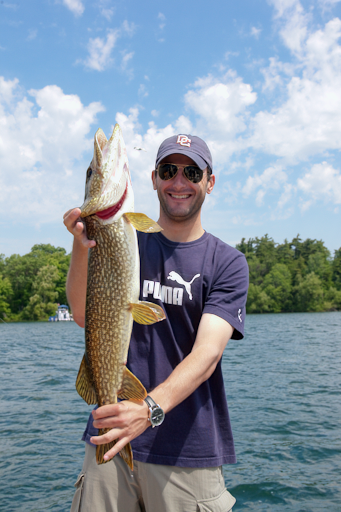}}
            \centerline{(b) DPE}\medskip
        \end{minipage}
        \hfill
        \begin{minipage}[b]{0.32\linewidth}
            \centering
            \centerline{\includegraphics[height=3cm, frame]{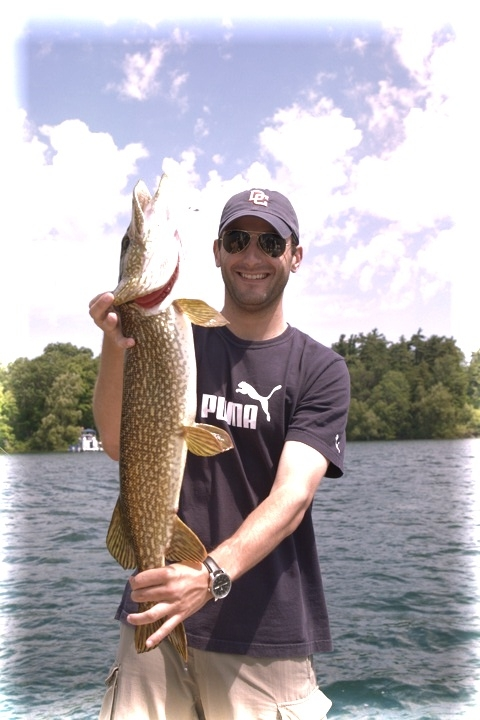}}
            \centerline{(c) UPE}\medskip
        \end{minipage}
    \end{minipage}
    \begin{minipage}[b]{1.0\linewidth}
        \begin{minipage}[b]{.32\linewidth}
            \centering
            \centerline{\includegraphics[height=3cm, frame]{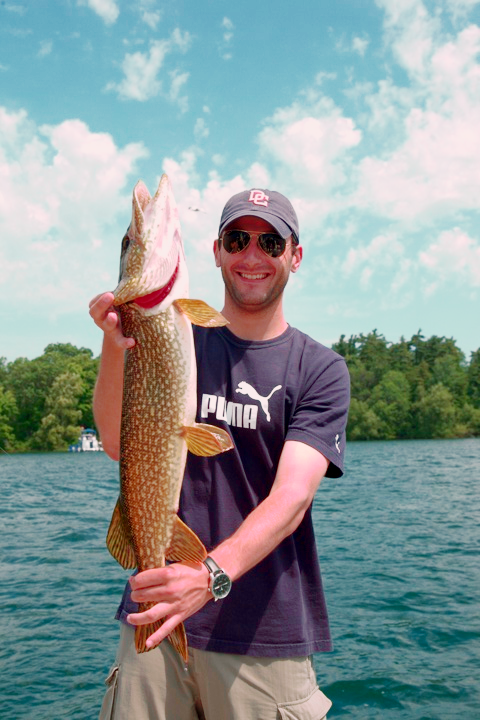}}
            \centerline{(d) CSRNet}\medskip
        \end{minipage}
        \hfill
        \begin{minipage}[b]{0.32\linewidth}
            \centering
            \centerline{\includegraphics[height=3cm, frame]{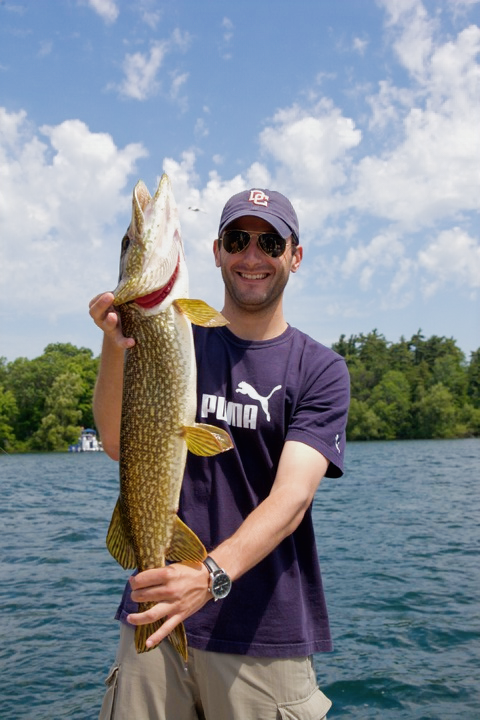}}
            \centerline{(e) Ours}\medskip
        \end{minipage}
        \hfill
        \begin{minipage}[b]{0.32\linewidth}
            \centering
            \centerline{\includegraphics[height=3cm, frame]{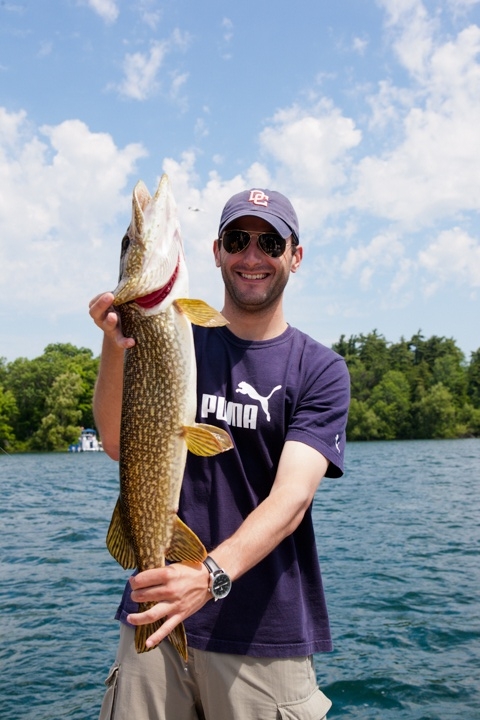}}
            \centerline{(f) Target}\medskip
        \end{minipage}
    \end{minipage}
    \caption{
    This photograph contains input image~(a) and target image~(f). In terms of white balance and exposure, the results of DPE~(b), UPE~(c), CSRNet~(d) are inconsistent with the ground truth.
    Our enhanced image~(e) is closer to the target.
    }
    \label{fig:intro}
\end{figure}

These difficulties encourage the creation of completely automated image enhancing strategies that may replace the labor of non-expert users or give experienced artists with a better manual editing starting point. Photographers commonly use a mix of local filters and global modifications to retouch pictures. With the advent of deep learning, automatic image enhancement optimizes images with neural network. There has appeared a lot of great works such as~\cite{moran2020deeplpf,he2020conditional,yang2022seplut,zeng2020learning,zhang2021star}

Despite the aforementioned works, there remain a few issues. First and foremost, some works may concentrate only on the pixel or global level. Consequently, they are incapable of capturing either the global tone or the fine-grained changes.
Second, few studies optimize a picture with diverse frequency priors, which may result in an image that is not entirely optimized and yields unsatisfactory results.

Inspired by wavelet transformation and vision transformer~(ViT)~\cite{dosovitskiy2020vit}, our objective is to extract additional features and enhance photographs in various frequency domains. We summarize our major contributions as follows: 

\begin{figure*}[ht]
    \begin{minipage}[b]{1.0\linewidth}
        \includegraphics[width=\linewidth]{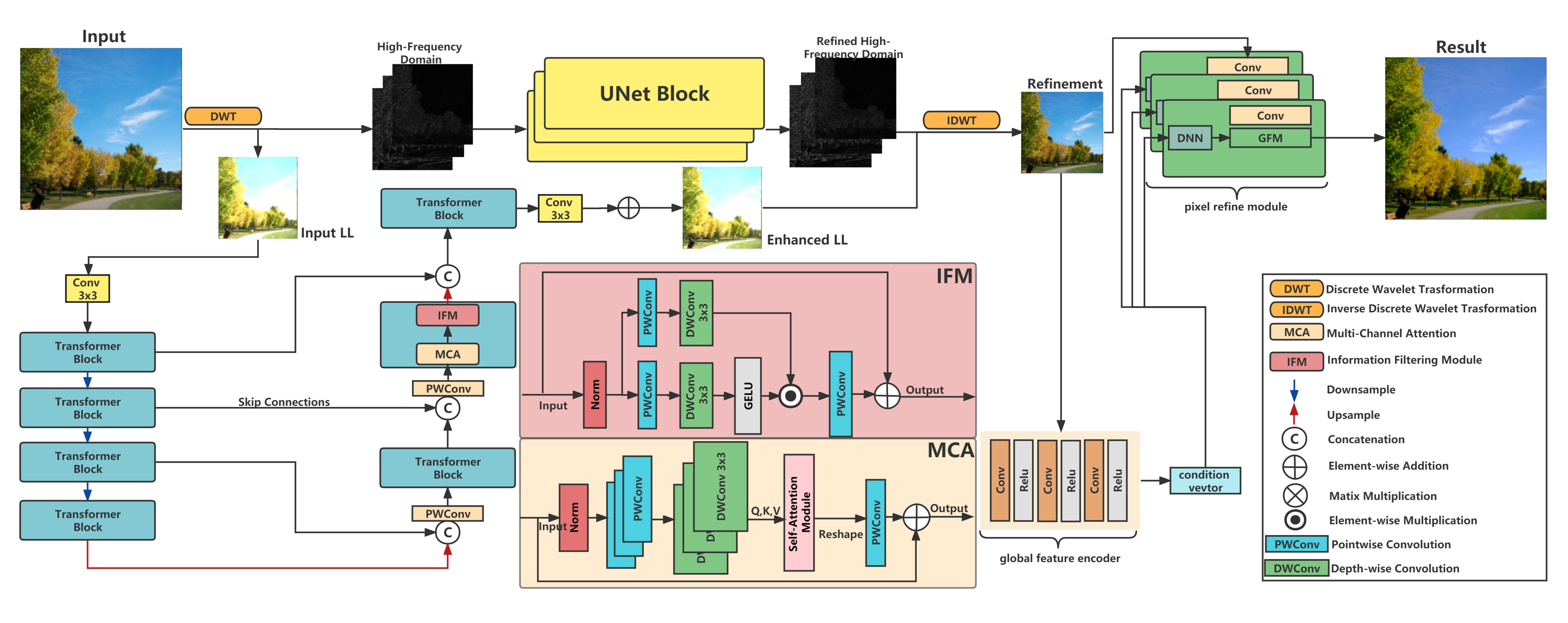}
    \end{minipage}
    \caption{
    The overall architecture of our WavEnhancer. It is comprised of three components: wavelet transform, global stylization remapping module for global feature restoration, and detailed parametric refinement module for visual quality improvement.
    }
    \label{fig:filmnet}
\end{figure*}

\begin{enumerate}
    \item We present the WavEnhancer, a novel wavelet domain transformer-based image enhancement framework.
    \item We demonstrate our model is superior to the state-of-the-art methods via experiments on publicly accessible benchmark datasets.
\end{enumerate}

\section{Related Work}

Recent years have witnessed the emergence of the deep learning model as a formidable competitor in the image enhancement task.

Some works concentrate on local image enhancement. UPE~\cite{wang2019underexposed} discovers a scaling luminance map via an encoder-decoder arrangement. The intricacy of the learned mapping is highly dependent on the regularization approach applied. DPE~\cite{chen2018deep} provided intermediate lighting into the network to link the input with the predicted enhancement outcomes. HDRNet~\cite{gharbi2017deep} applied bilateral grid processing and local affine color transforms. Similar to HDRNet, a light-weight framework called CSRNet~\cite{he2020conditional} was constructed for efficient retouching. Later, DeepLPF~\cite{moran2020deeplpf} enhanced images using learned spatially local filters. 

Others focus on global image enhancement. Bychkovsky \emph{et al.}~\cite{bychkovsky2011learning} constructed the well-known MIT-Adobe-5K dataset. They also present a regression-based method for discovering the photographic modifications of artists from picture pairings. STAR-DCE~\cite{zhang2021star} proposed a lightweight transformer network for enhancing real-time image quality. 3D-LUT~\cite{zeng2020learning} proposed to learn 3D LUTs using annotated data. SepLUT~\cite{yang2022seplut} partitioned a single color transformation into component-independent and component-correlated subtransforms.

\section{METHODOLOGY}

\subsection{Overall framework}

Our model comprises wavelet transform~\cite{You2020FinePG}, a global stylization remapping module~(GSR) and a detailed parametric refinement module~(DPR). The entire workflow is shown in Figure~\ref{fig:filmnet}.

Initially, our model is fed a pair images, which is then subjected to Discrete Wavelet Transformation~(DWT). Each image is divided into discrete channels: LL is the low frequency region, while LH, HL and HH are the high frequency sectors. The low frequency region captures approximation coefficients such as geometric features and color context, whereas the high frequency sectors retrieve textural information.

The high-frequency sectors are subsequently sent to UNet blocks~\cite{ronneberger2015u} with Smooth L1 for convergence, whereas the low-frequency portion is processed by our global stylization remapping module. These two modules generate the correspondingly refined components. Then, Inverse Discrete Wavelet Transformation~(IDWT) integrates these outputs to reconstruct the image, and the result is further enhanced by our detailed parametric refinement module, ultimately producing a stylized output.

\subsection{Global stylization remapping}
After the DWT, the LL band is a thumbnail of the image content that retains the original image's content information. Inspired by Restormer~\cite{zamir2022restormer}, we apply transformer blocks for the stylization remapping of LL. The self-attention module of transformer can effectively gather global information, which is suitable for LL that requires color and texture refinement.

In the first stage, we apply a $3 \times 3$ convolution to the LL to extract its low-level features and the output shape is $\frac{H}{2} \times \frac{W}{2} \times C$. Then, these features are transformed into deep features by four symmetric encoder-decoder, i.e., the transformer blocks, where the number of blocks increasing from top to bottom to preserve efficiency. The up and down sampling between 4 blocks are achieved by pixel reshuffle. 

Note that the Information Filtering Module~(IFM) and Muti-Channel Attention~(MCA) are included in each transformer block. By computing the attention on the channel to implicitly encode the global contextual information, the MCA computes the self-attention on the channel as opposed to spatially. $Q (query)$, $K (key)$ and $V (value)$ are generated using depth-wise convolution operations in between the computation of self-attentive maps so that local information can be emphasized. In the meantime, by regulating the flow of information, the IFM at each level enables each level to concentrate on minute features that complement the other levels. In other words, IFM performs a distinct function compared to the MCA. 

Finally, the LL part is recovered by $3 \times 3$ convolution, with the shape of $\frac{H}{2} \times \frac{W}{2} \times C$. Given a set of N images pairs $\left\{\left(Y_i, \hat{Y}_i\right)\right\}_{i=1}^N$, where $Y_i$ is the ground truth LL and $\hat{Y}_i$ is the enhanced LL. The refinement loss $L_{R}$ is defined as Equation~\ref{eqn:loss}:
\begin{equation}
\label{eqn:loss}
\begin{aligned}
L_{R} &=\sum_{i=1}^N\left\{\omega_{\text {lab }}\left\|\operatorname{Lab}\left(\hat{Y}_i\right)-\operatorname{Lab}\left(Y_i\right)\right\|_1\right.\\
&\left.+\omega_{\text {ms-ssim}} \operatorname{MS-SSIM}\left(L\left(\hat{Y}_i\right), L\left(Y_i\right)\right)\right\}
\end{aligned}
\end{equation}
in which $Lab(x)$ returns CIELab channels corresponding to the original images' RGB channels, and $L(x)$ returns the image's CIELab L channel. MS-SSIM is the multi-scale structural similarity, and $\omega_{\text {lab }}$, $\omega_{\text {ms-ssim}}$ are hyperparameters that weight the relative importance of components in the loss function.

Next, the enhanced LL and refined high-frequency domain are merged by IDWT, which then outputs a refined intermediate result.

\subsection{Detailed parametric refinement}

Inspired by CSRNet~\cite{he2020conditional}, at the following stage, the intermediate result from previous part is sent to our Detailed Parametric Refinement Module for better performance. Inside the DPR, the pixel refine module receives the low-quality image as input and generates the stylized image, while global feature encoder estimates the priors from the input image and then modulates the former one by Global Feature Modulation~(GFM) operations. Here we also use Equation 1 to converge the model.

The pixel refine module is a fully convolutional architecture with $N$ layers and $N \times 1$ ReLU activations. The unique part of pixel refine module is the fact that all filter sizes are $1 \times 1$, indicating that each pixel in the input image is separately altered. Consequently, the pixel refine module operates independently on each pixel and glides over the input image. To obtain the global information, three blocks are contained inside the global feature encoder: convolutional, ReLU, and downsampling layers. The output of the global feature encoder is a condition vector that will be sent to the pixel refine module. 

GFM is actually a special AdaFM~\cite{he2019modulating}, which can be represented by: 
$AdaFM(x_i)=g_i \times {x_i} + b_i$, where $g_i$ and $b_i$ are the filter and bias. When the filter $g_i$ is of size $1 \times 1$, AdaFM becomes GFM. In another perspective, GFM is also able to scale and shift the feature map $x_i$ without normalizing it when provided with affine parameters $\gamma,\beta$ as shown in Equation~\ref{eqn:GFM}:
\begin{equation}
\label{eqn:GFM}
GFM(x_i)=\gamma\times{x_i}+\beta
\end{equation}

\subsection{Criteria and implementation}

Taking into account the feature-level information in various contexts, We train our network with perceptual loss~\cite{zhang2018single,johnson2016perceptual} $L_{\Phi}$ in the feature domain, as follows:

\begin{equation}
\label{eqn:perc}
L_{\Phi} = \sum^{5}_{k=0}\lambda_{l}||\Phi_{k}( I_{free}^{'})) - \Phi_{k}(I_{free}) ||_1,
\end{equation}
where $\Phi$ is the ImageNet-pretrained VGG16 model. We assess the differences between the CONVk2~($k = 1...5$) and the original pictures~($k = 0$ in Equation~\ref{eqn:perc}) between the ground truth and the enhanced image. In addition, our model for pixel-wise enhancement is constrained by Smooth L1 loss $L_{1;smooth}$.

By incorporating all regularization items, the final loss function of our model is as follows:

\begin{equation}
\label{eqn:total}
L_{total} = L_{\Phi} + \lambda_{R} L_{R} + \lambda_{1;smooth} L_{1;smooth}
\end{equation}
where the two constant parameters $\lambda_{R}$ and $\lambda_{1;smooth}$ are used to control the effects of the feature and pixel regularization terms respectively. Emperically, we set $\lambda_{R} = 2$ and $\lambda_{1;smooth} = 2$.

Our implementation is based on PyTorch. In order to train our model, the standard Adam optimizer with default settings is employed. We set the batch size to 1 and the learning rate to $1e-4$. We utilize random cropping, horizontal flipping, and brightness and saturation adjustments to enable data augmentation.

\begin{table*}[!ht]
\centering
\resizebox{0.65\linewidth}{!}{
\begin{tabular}{c|ccc|ccc}
\hline
                         & \multicolumn{3}{c|}{FiveK}                                                                                           & \multicolumn{3}{c}{HDR+}                                                                                             \\ \cline{2-7} 
\multirow{-2}{*}{Method} & PSNR $\uparrow$                       & SSIM $\uparrow$                       & $\Delta E$ $\downarrow$              & PSNR $\uparrow$                       & SSIM $\uparrow$                       & $\Delta E$ $\downarrow$              \\ \hline
HDRNet                   & 19.93                                 & 0.798                                 & 14.42                                & 23.04                                 & 0.879                                 & 8.97                                 \\
DPE                      & 17.66                                 & 0.725                                 & 17.71                                & 22.56                                 & 0.872                                 & 10.45                                \\
UPE                      & 21.88                                 & 0.853                                 & 10.80                                & 21.21                                 & 0.816                                 & 13.05                                \\
CSRNet                   & 17.85                                 & 0.790                                 & 18.27                                & N/A                                   & N/A                                   & N/A                                  \\
DeepLPF                  & 24.55                                 & 0.846                                 & 8.62                                 & N/A                                   & N/A                                   & N/A                                  \\
3D-LUT                   & 24.59                                 & 0.846                                 & 8.30                                 & 23.54                                 & 0.885                                 & 7.93                                 \\
STAR-DCE                 & 24.5                                  & 0.893                                 & N/A                                  & 26.5                                  & 0.883                                 & 5.77                                 \\
SepLUT                   & 25.02                                 & 0.873                                 & 7.91                                 & N/A                                   & N/A                                   & N/A                                  \\ \hline
GSR+UNet                 & 24.28                                 & 0.874                                 & 8.57                                 & 24.17                                 & 0.823                                 & 8.89                                 \\
ConvNet+DPR              & 24.40                                 & 0.873                                 & 8.43                                 & 26.84                                 & 0.867                                 & 6.34                                 \\
ConvNet+UNet             & 21.43                                 & 0.815                                 & 12.23                                & 25.85                                 & 0.860                                 & 7.28                                 \\
UNet+DPR                 & 24.06                                 & 0.879                                 & 9.42                                 & 25.48                                 & 0.862                                 & 7.87                                 \\
UNet+UNet                & 22.04                                 & 0.838                                 & 11.72                                & 23.87                                 & 0.840                                 & 9.49                                 \\
Ours                     & {\color[HTML]{FF0000} \textbf{25.46}} & {\color[HTML]{FF0000} \textbf{0.896}} & {\color[HTML]{FF0000} \textbf{7.28}} & {\color[HTML]{FF0000} \textbf{28.68}} & {\color[HTML]{FF0000} \textbf{0.905}} & {\color[HTML]{FF0000} \textbf{4.89}} \\ \hline
\end{tabular}
}
\caption{Quantitative comparisons on the MIT FiveK and HDR+ dataset of different image enhancement methods. "N/A" indicates that the result is unavailable. The top result is highlighted in red.}
\label{table:exp_1}
\end{table*}

\begin{figure*}[!ht]
    \begin{minipage}[b]{1.0\linewidth}
        \begin{minipage}[b]{0.12\linewidth}
            \centering
            \centerline{\includegraphics[height=4.35cm, frame]{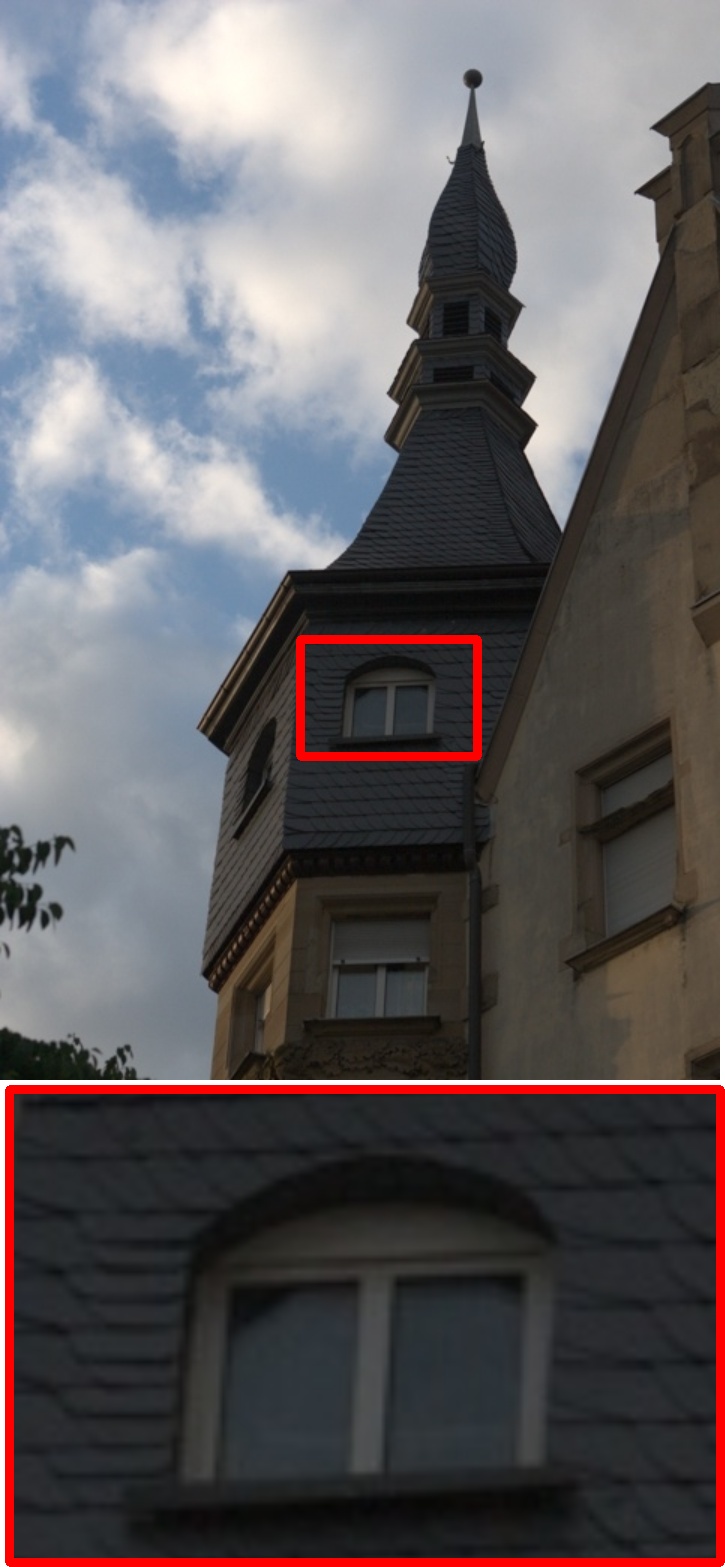}}
        \end{minipage}
        \begin{minipage}[b]{.12\linewidth}
            \centering
            \centerline{\includegraphics[height=4.35cm, frame]{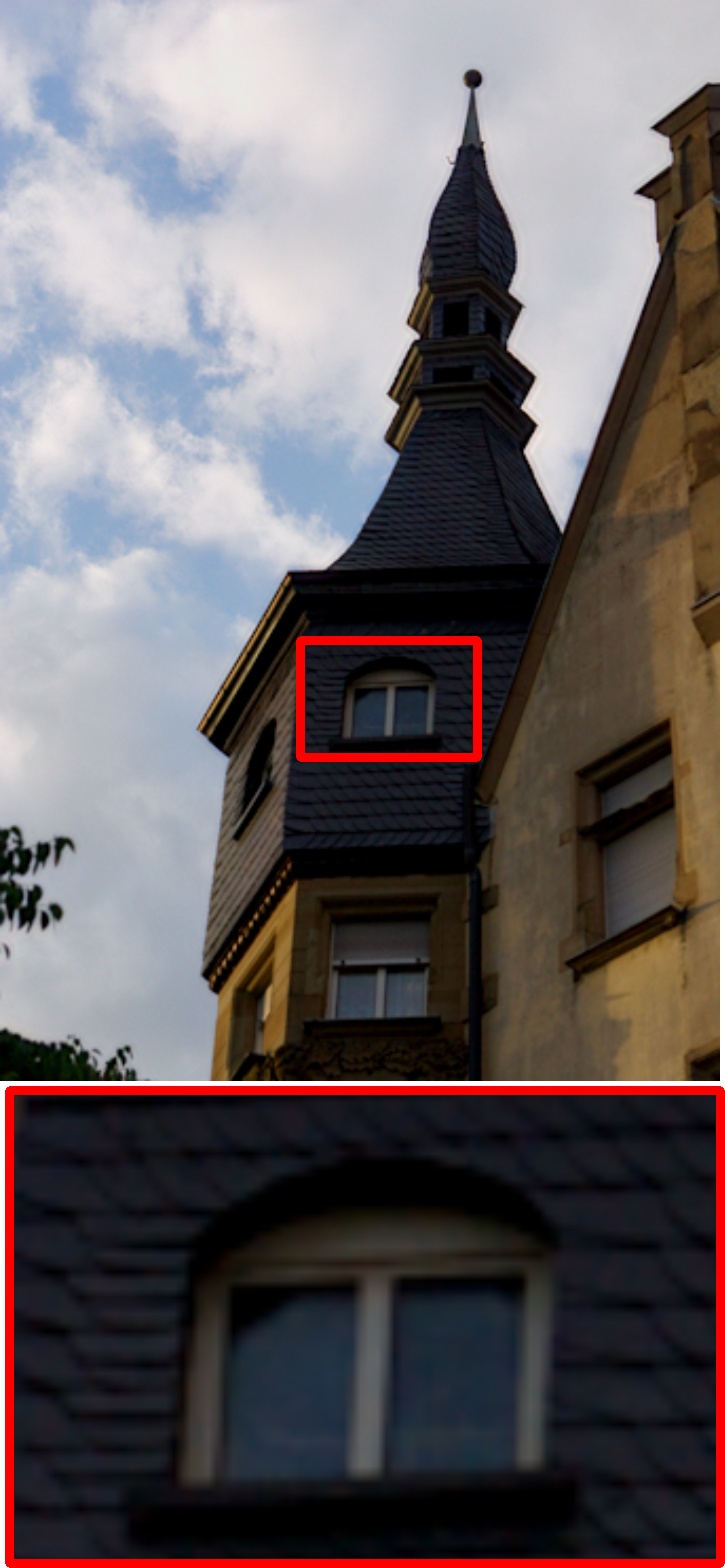}}
        \end{minipage}
        \begin{minipage}[b]{0.12\linewidth}
            \centering
            \centerline{\includegraphics[height=4.35cm, frame]{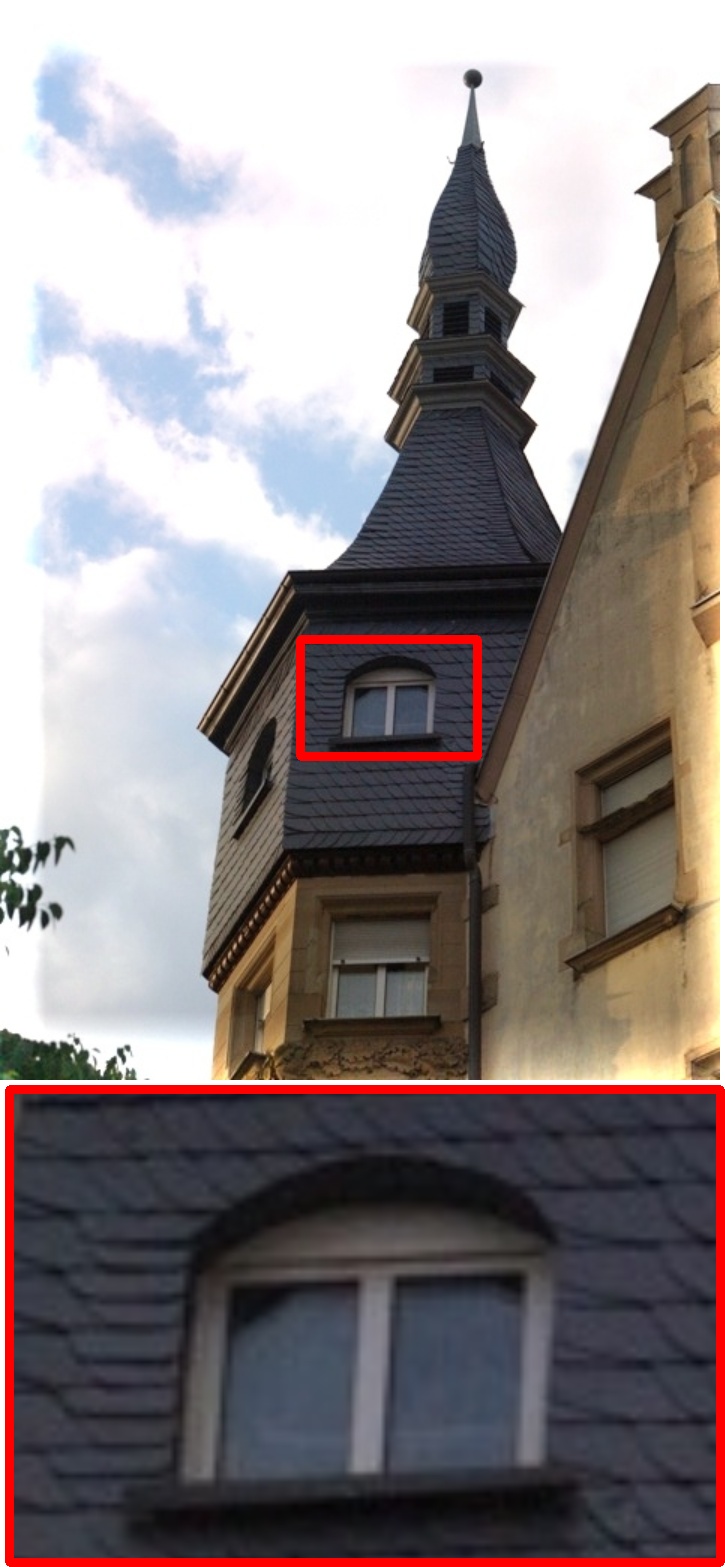}}
        \end{minipage}
        \begin{minipage}[b]{0.12\linewidth}
            \centering
            \centerline{\includegraphics[height=4.35cm, frame]{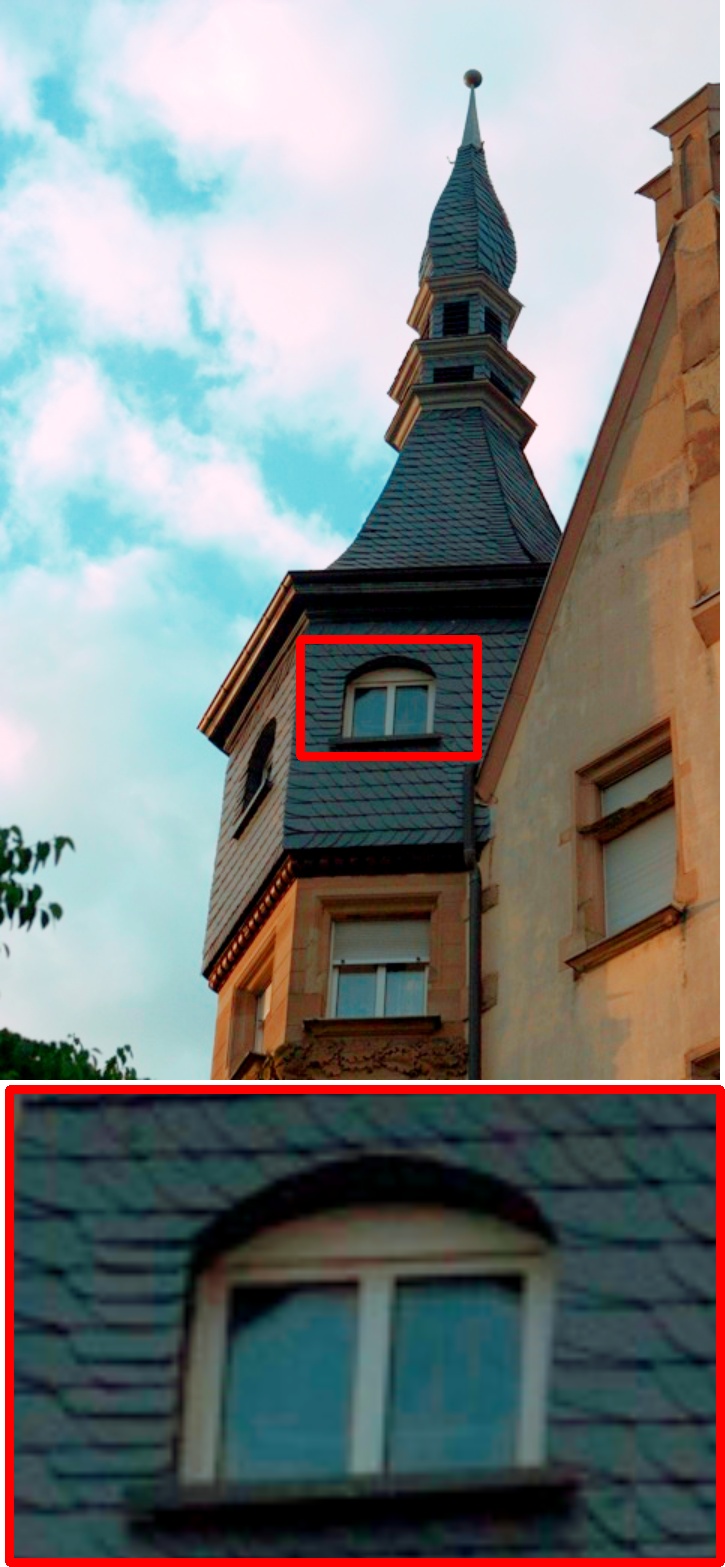}}
        \end{minipage}
        \begin{minipage}[b]{.12\linewidth}
            \centering
            \centerline{\includegraphics[height=4.35cm, frame]{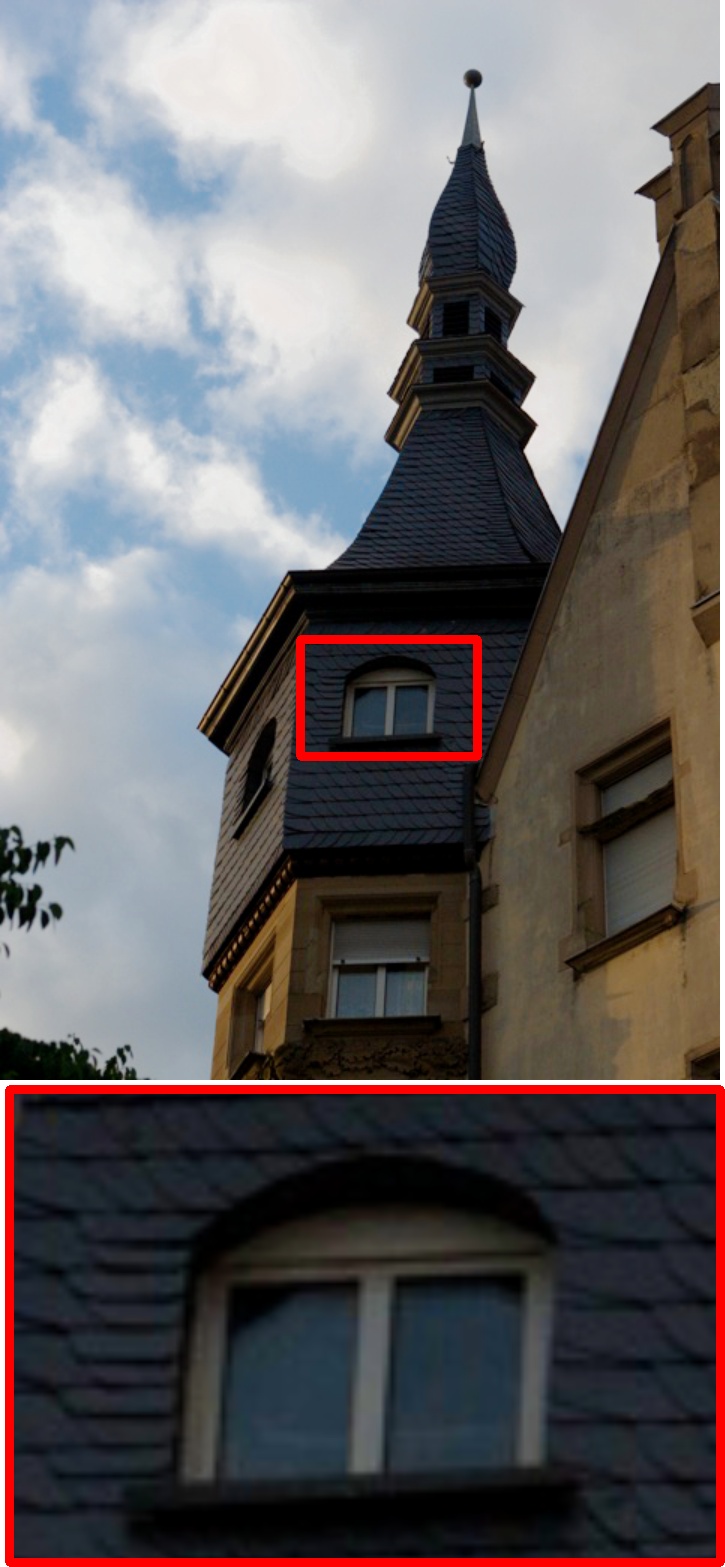}}
        \end{minipage}
        \begin{minipage}[b]{0.12\linewidth}
            \centering
            \centerline{\includegraphics[height=4.35cm, frame]{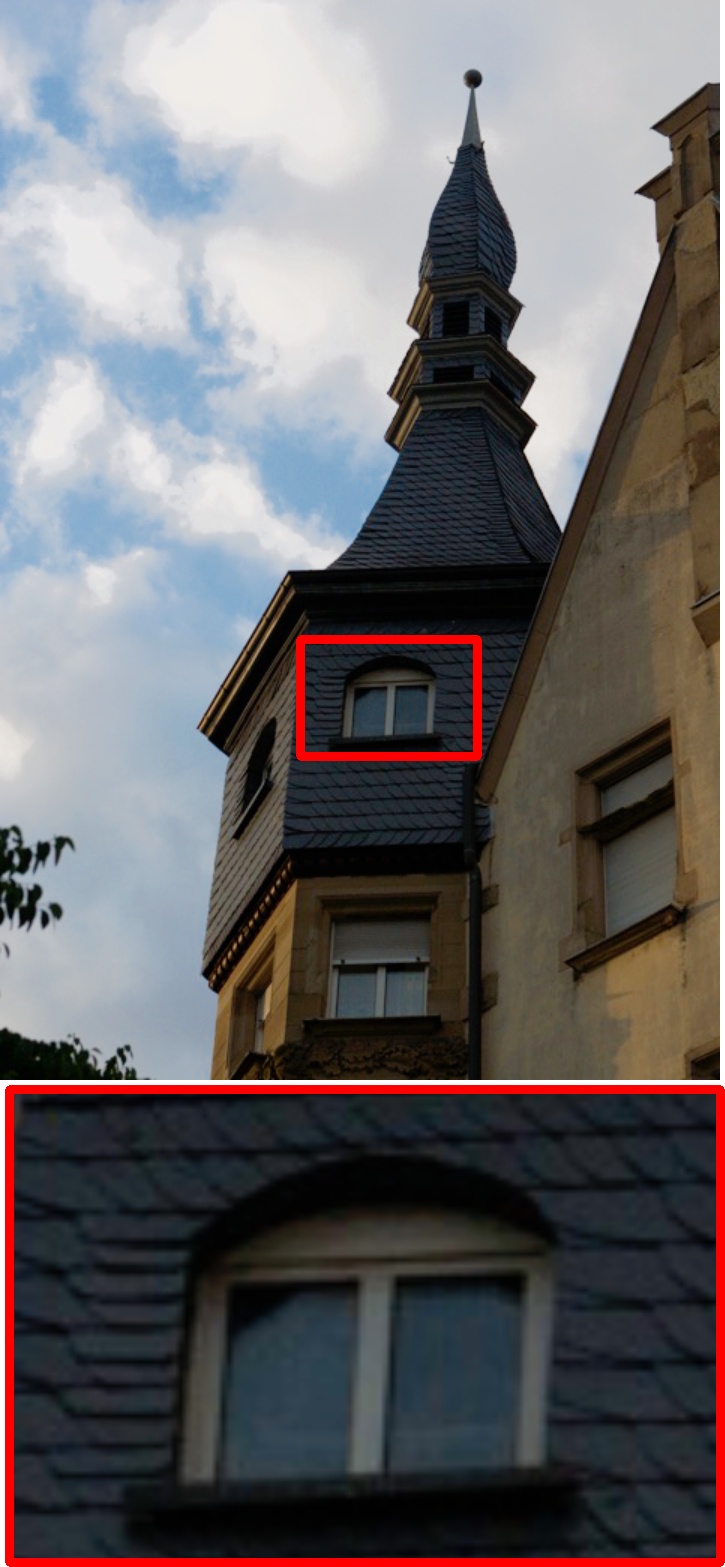}}
        \end{minipage}
        \begin{minipage}[b]{0.12\linewidth}
            \centering
            \centerline{\includegraphics[height=4.35cm, frame]{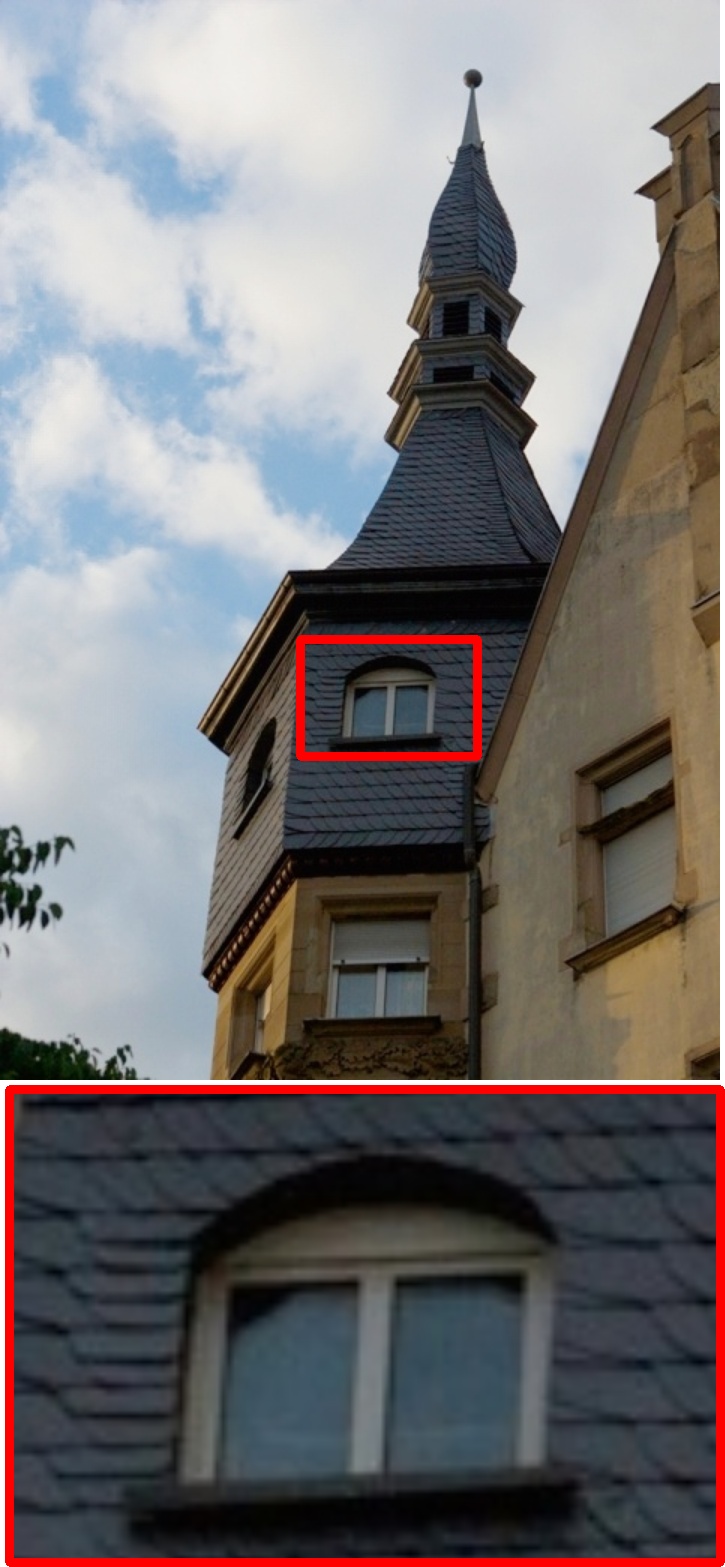}}
        \end{minipage}
        \begin{minipage}[b]{0.12\linewidth}
            \centering
            \centerline{\includegraphics[height=4.35cm, frame]{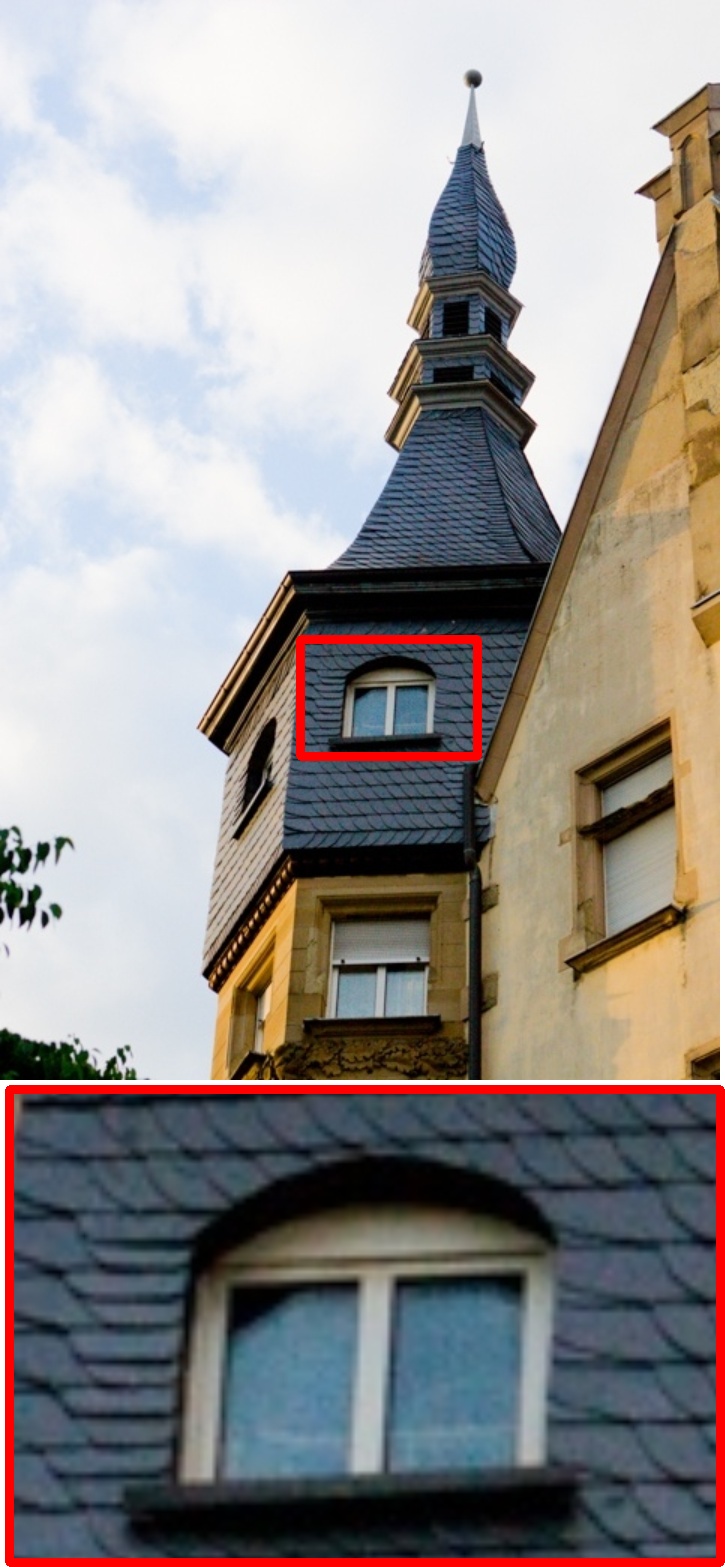}}
        \end{minipage}
    \end{minipage}

    \begin{minipage}[b]{1.0\linewidth}
        \begin{minipage}[b]{0.12\linewidth}
            \centering
            \centerline{\includegraphics[height=4.35cm, frame]{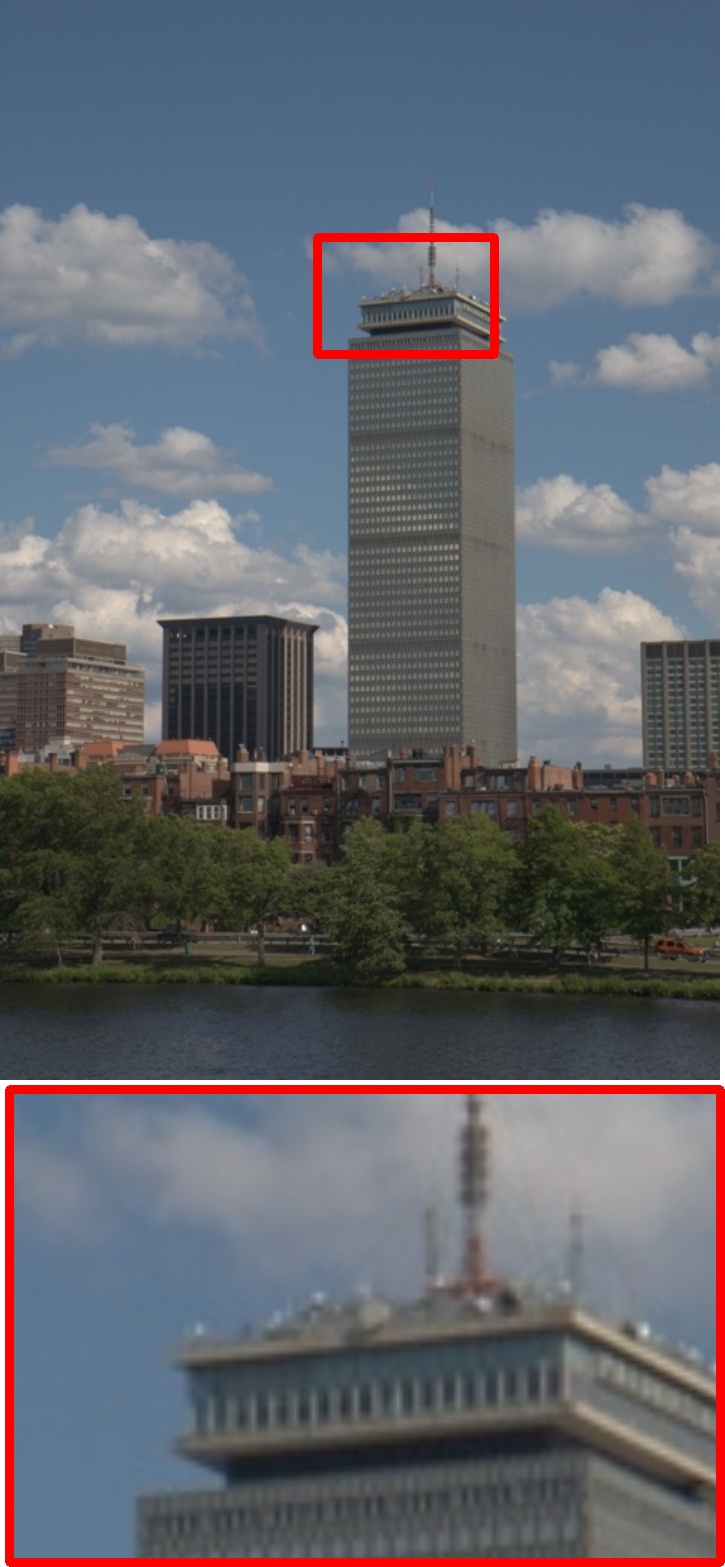}}
        \end{minipage}
        \begin{minipage}[b]{.12\linewidth}
            \centering
            \centerline{\includegraphics[height=4.35cm, frame]{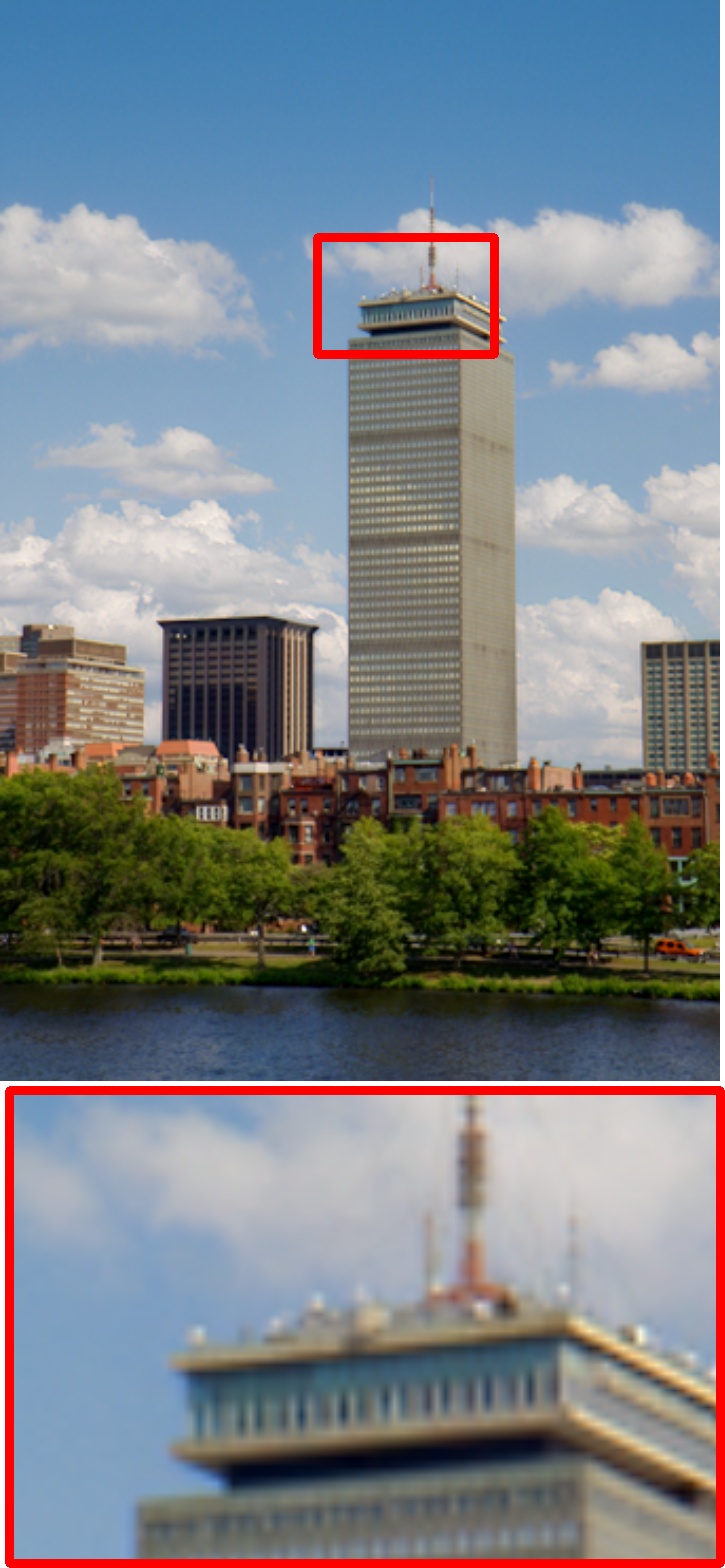}}
        \end{minipage}
        \begin{minipage}[b]{0.12\linewidth}
            \centering
            \centerline{\includegraphics[height=4.35cm, frame]{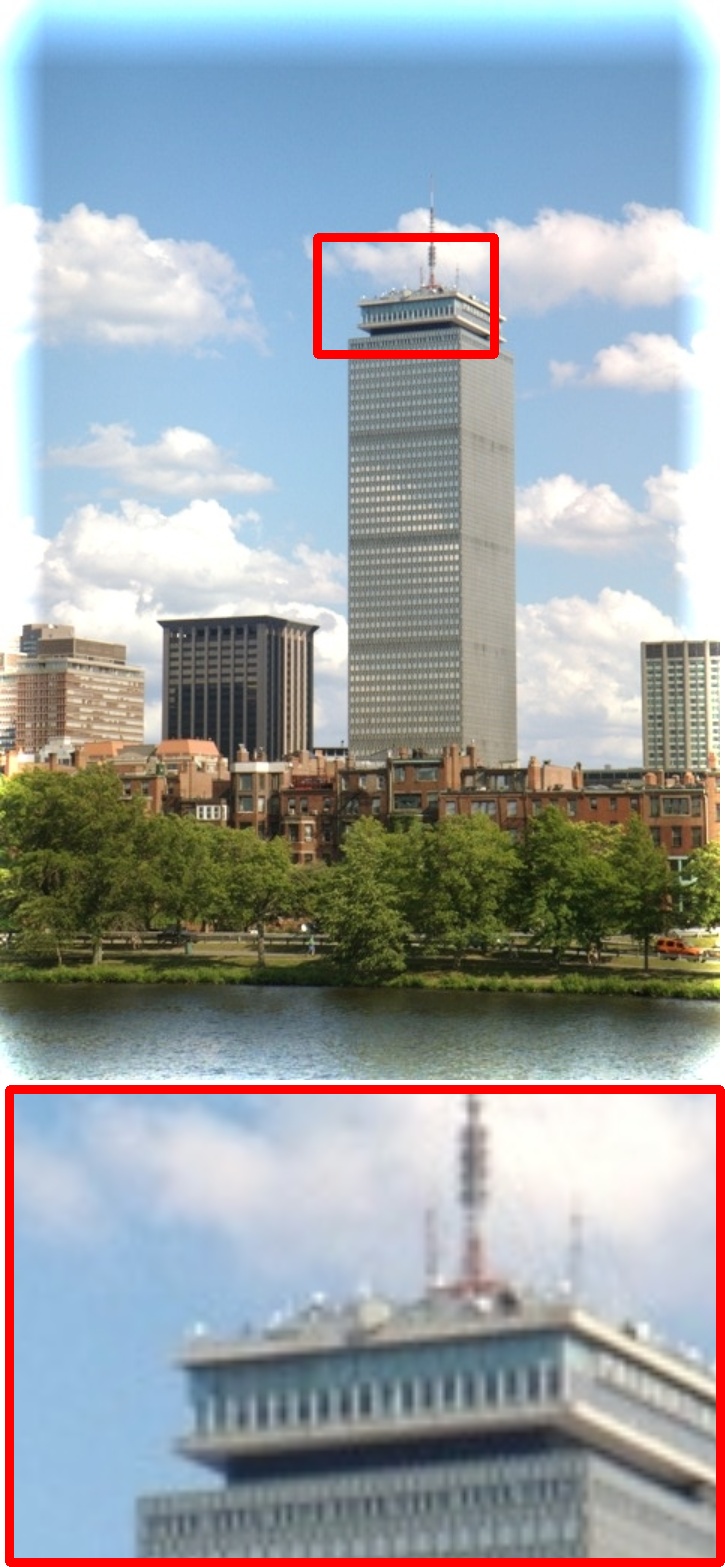}}
        \end{minipage}
        \begin{minipage}[b]{0.12\linewidth}
            \centering
            \centerline{\includegraphics[height=4.35cm, frame]{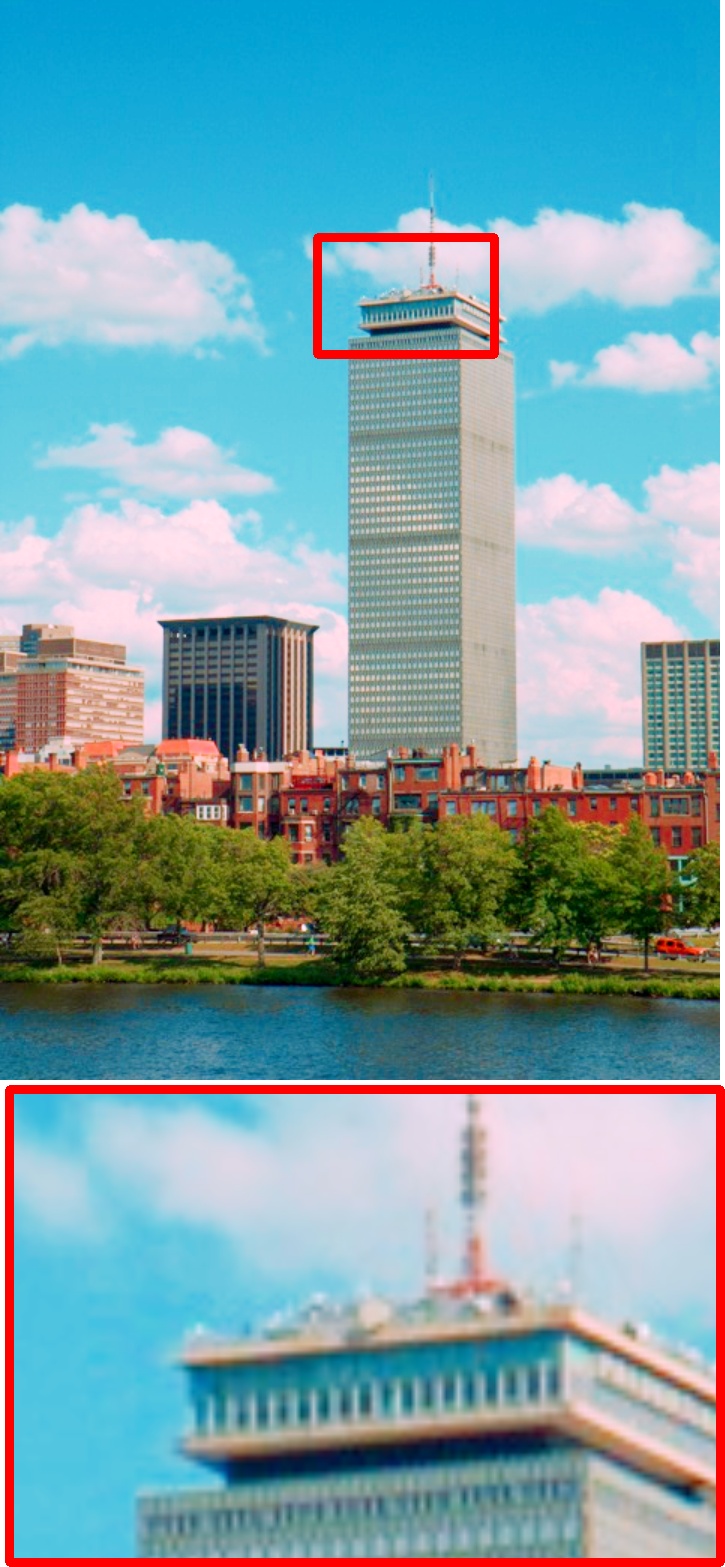}}
        \end{minipage}
        \begin{minipage}[b]{.12\linewidth}
            \centering
            \centerline{\includegraphics[height=4.35cm, frame]{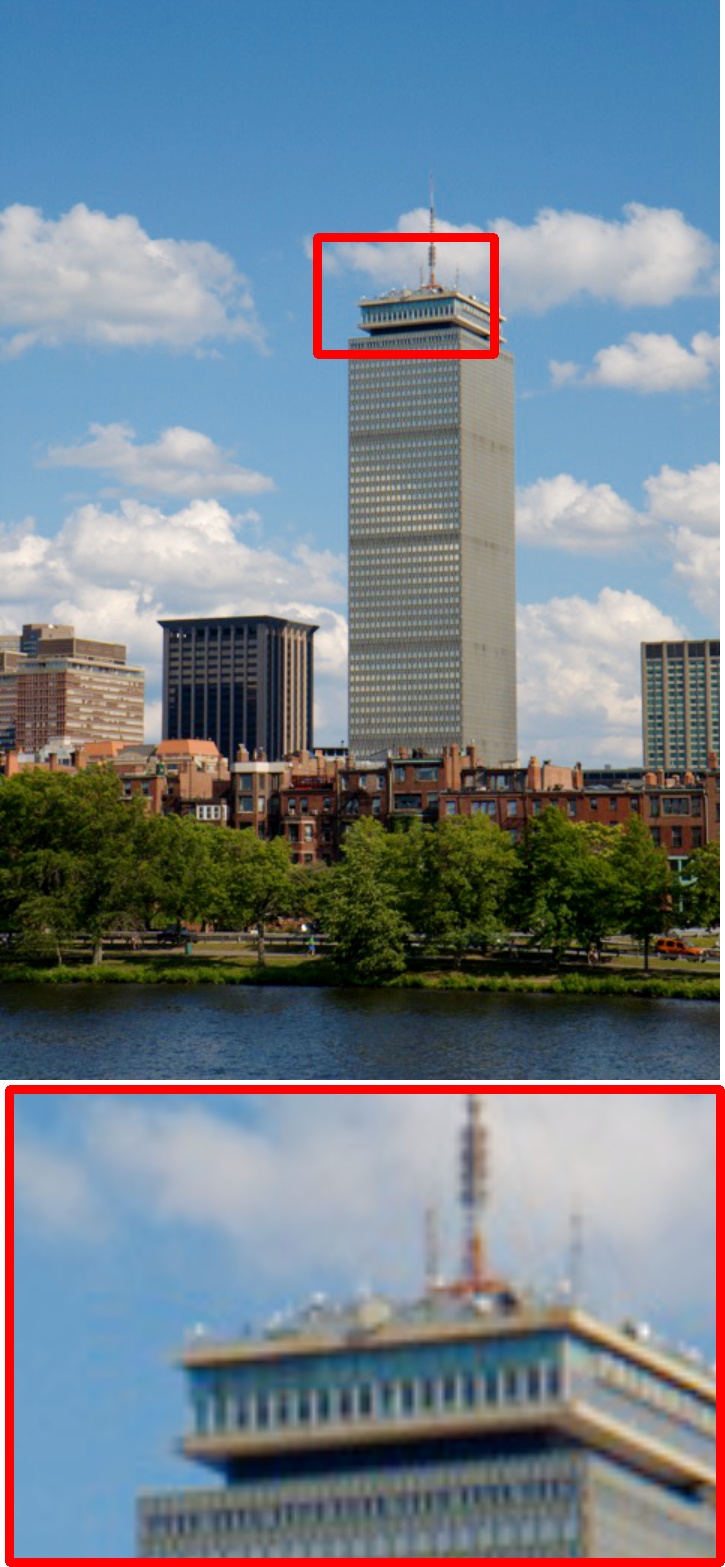}}
        \end{minipage}
        \begin{minipage}[b]{0.12\linewidth}
            \centering
            \centerline{\includegraphics[height=4.35cm, frame]{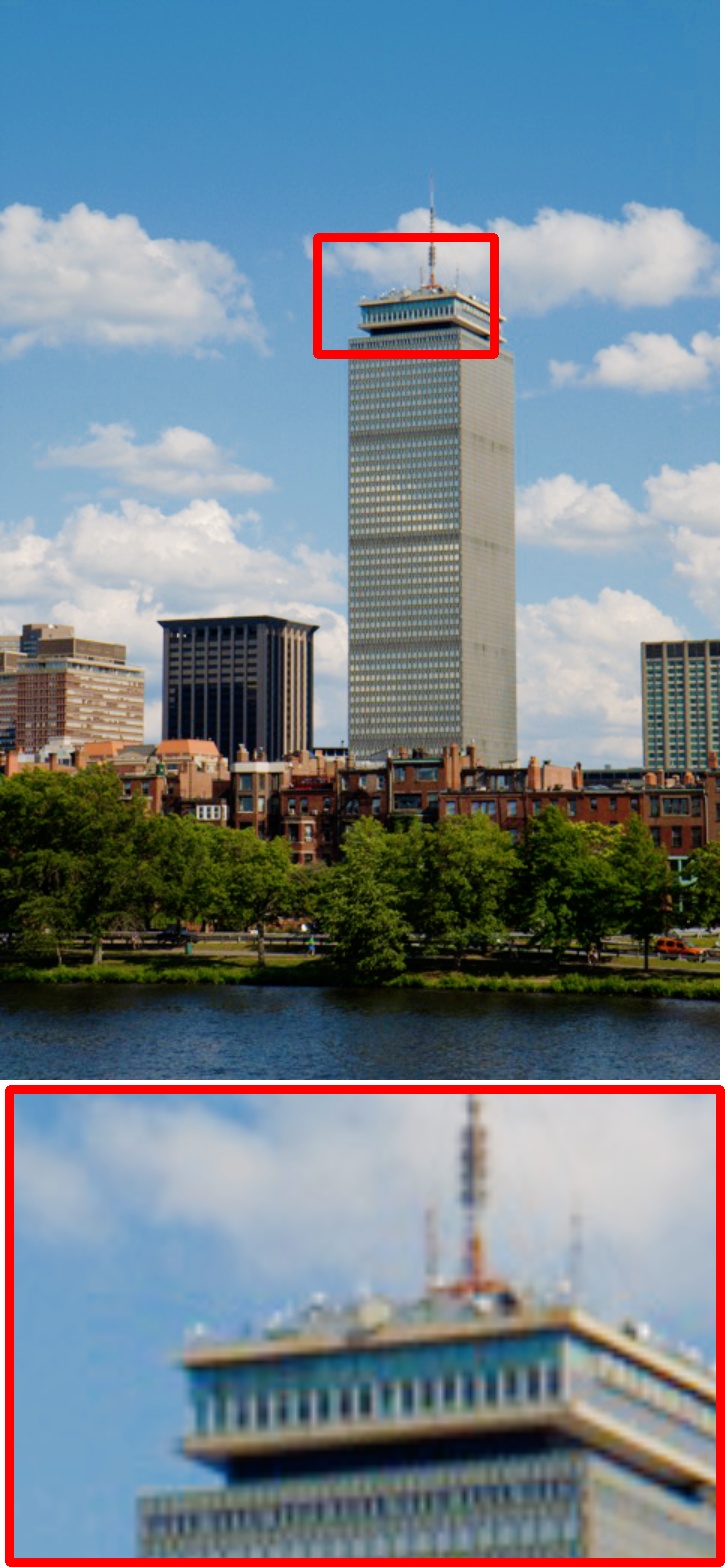}}
        \end{minipage}
        \begin{minipage}[b]{0.12\linewidth}
            \centering
            \centerline{\includegraphics[height=4.35cm, frame]{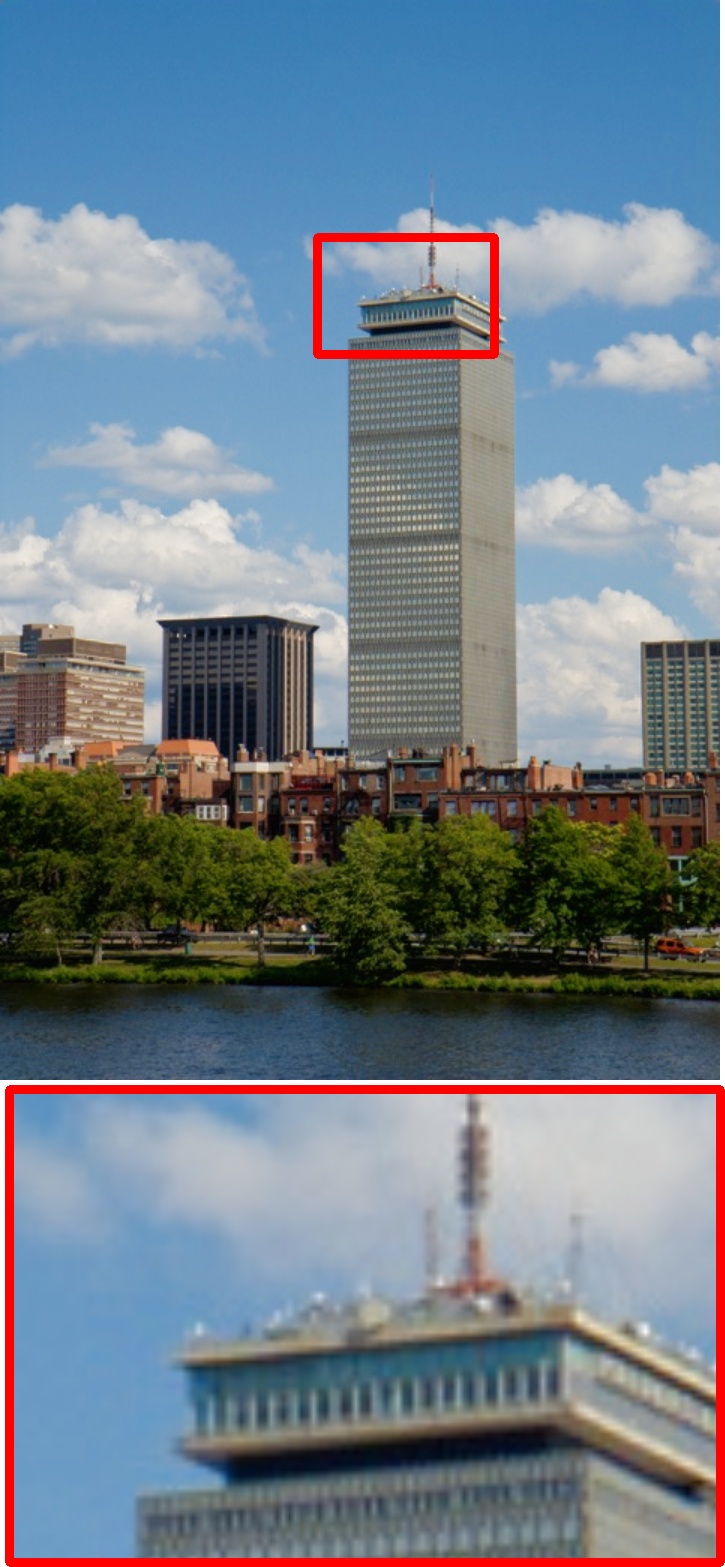}}
        \end{minipage}
        \begin{minipage}[b]{0.12\linewidth}
            \centering
            \centerline{\includegraphics[height=4.35cm, frame]{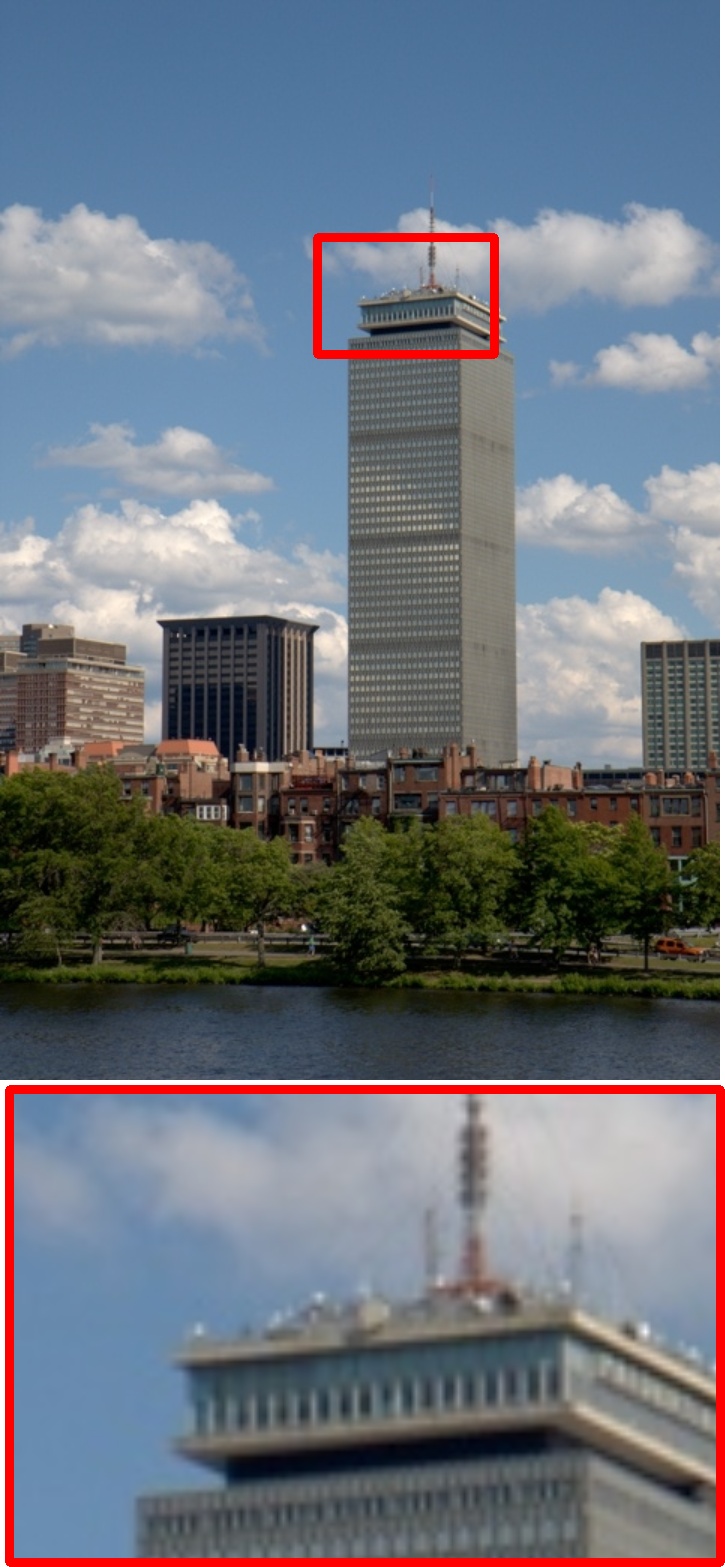}}
        \end{minipage}
    \end{minipage}
    
    \begin{minipage}[b]{1.0\linewidth}
        \begin{minipage}[b]{0.12\linewidth}
            \centering
            \centerline{\includegraphics[height=4.35cm, frame]{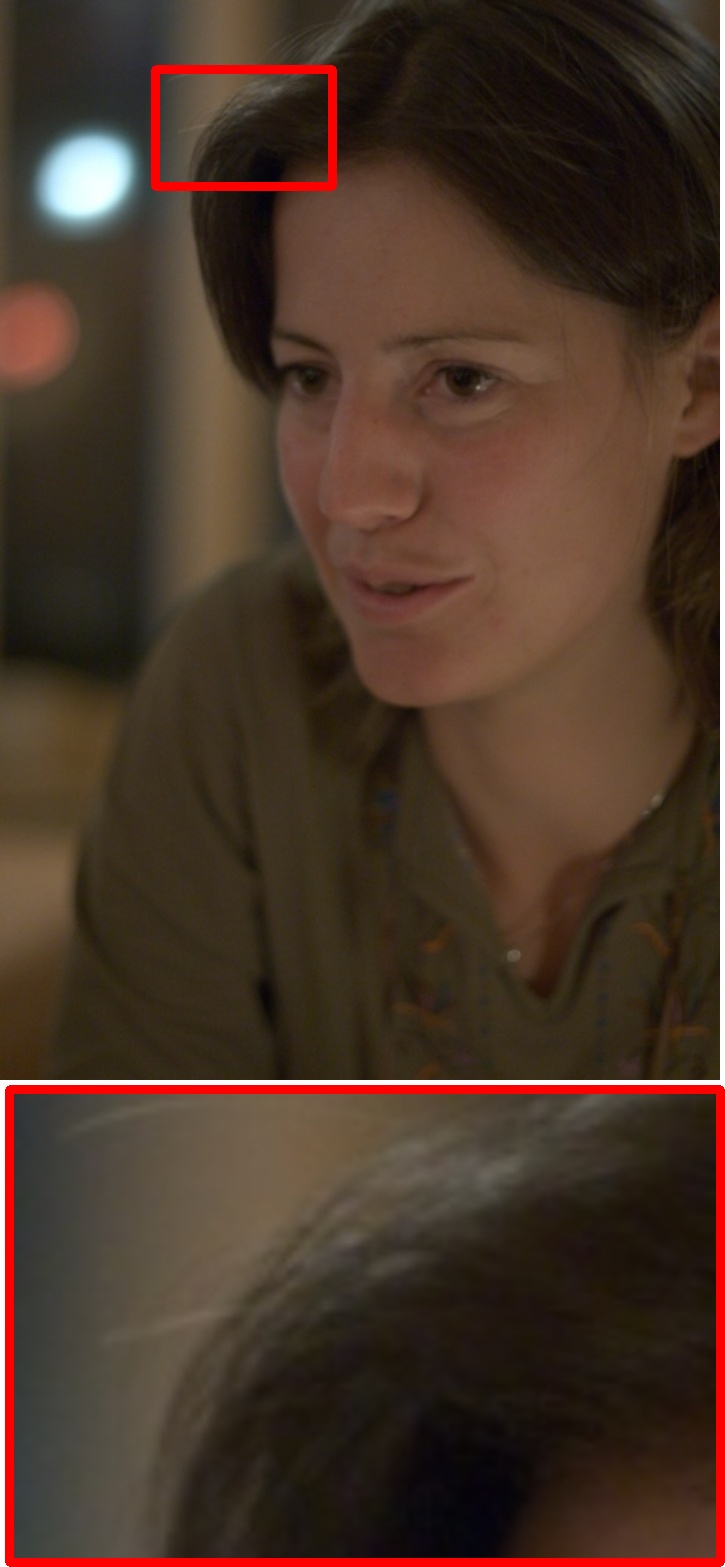}}
            \centerline{(a) Input}\medskip
        \end{minipage}
        \begin{minipage}[b]{.12\linewidth}
            \centering
            \centerline{\includegraphics[height=4.35cm, frame]{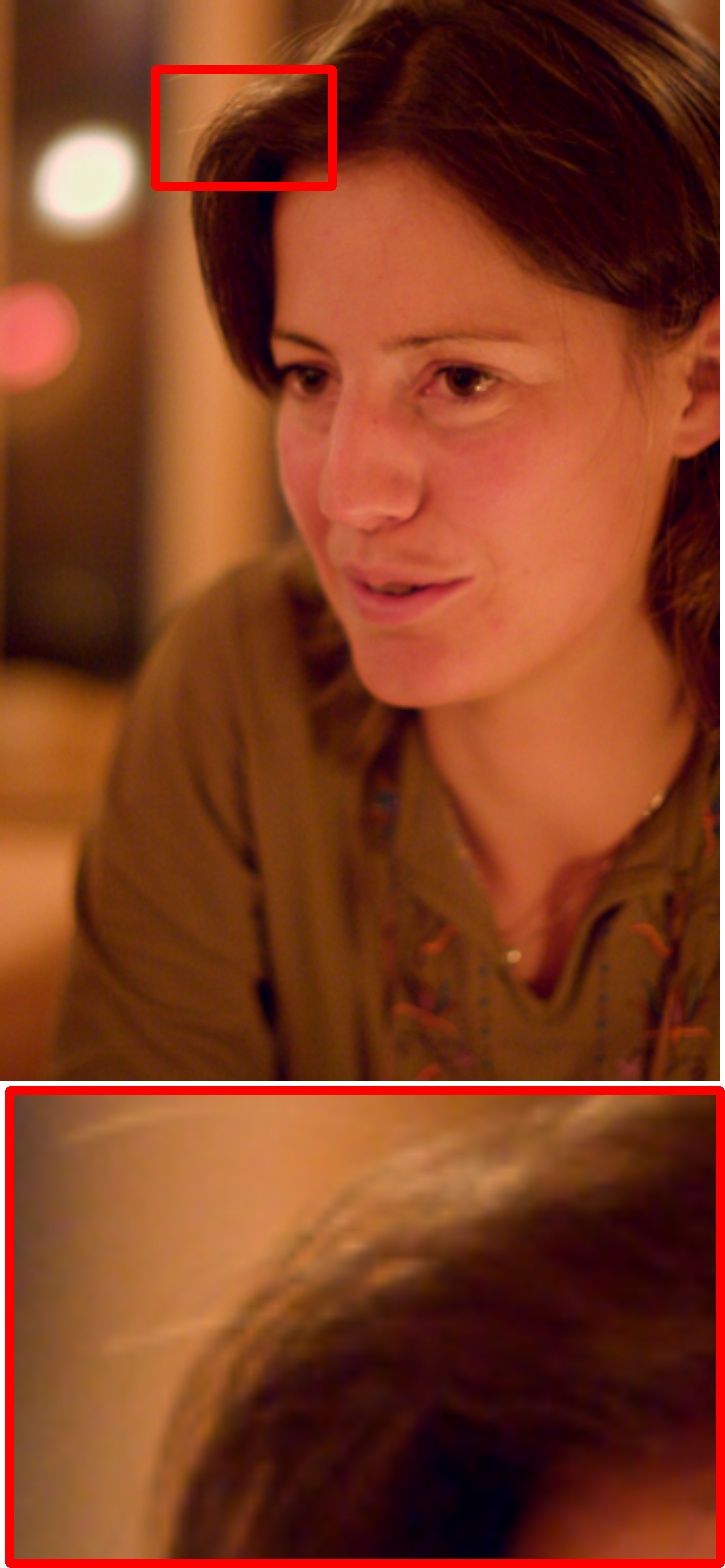}}
            \centerline{(b) DPE}\medskip
        \end{minipage}
        \begin{minipage}[b]{0.12\linewidth}
            \centering
            \centerline{\includegraphics[height=4.35cm, frame]{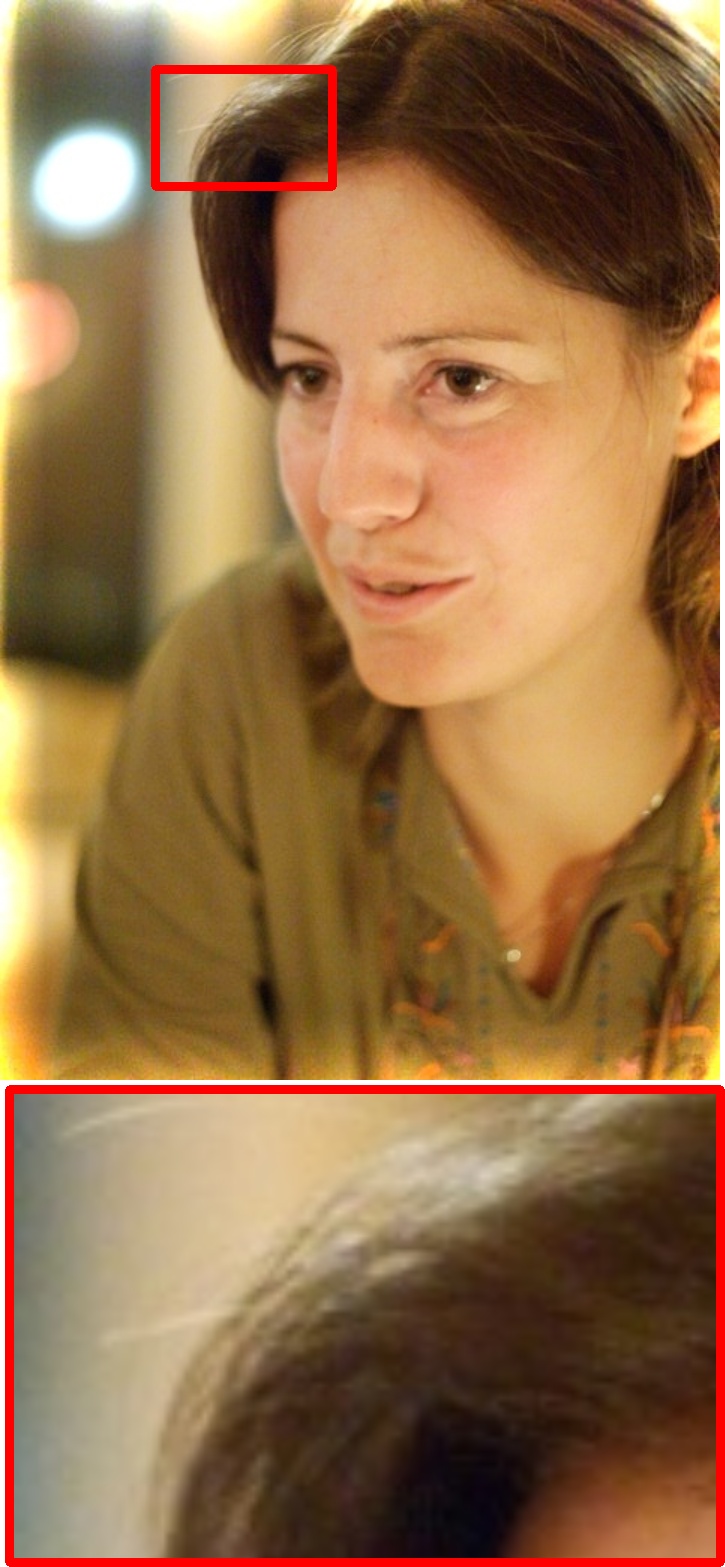}}
            \centerline{(c) UPE}\medskip
        \end{minipage}
        \begin{minipage}[b]{0.12\linewidth}
            \centering
            \centerline{\includegraphics[height=4.35cm, frame]{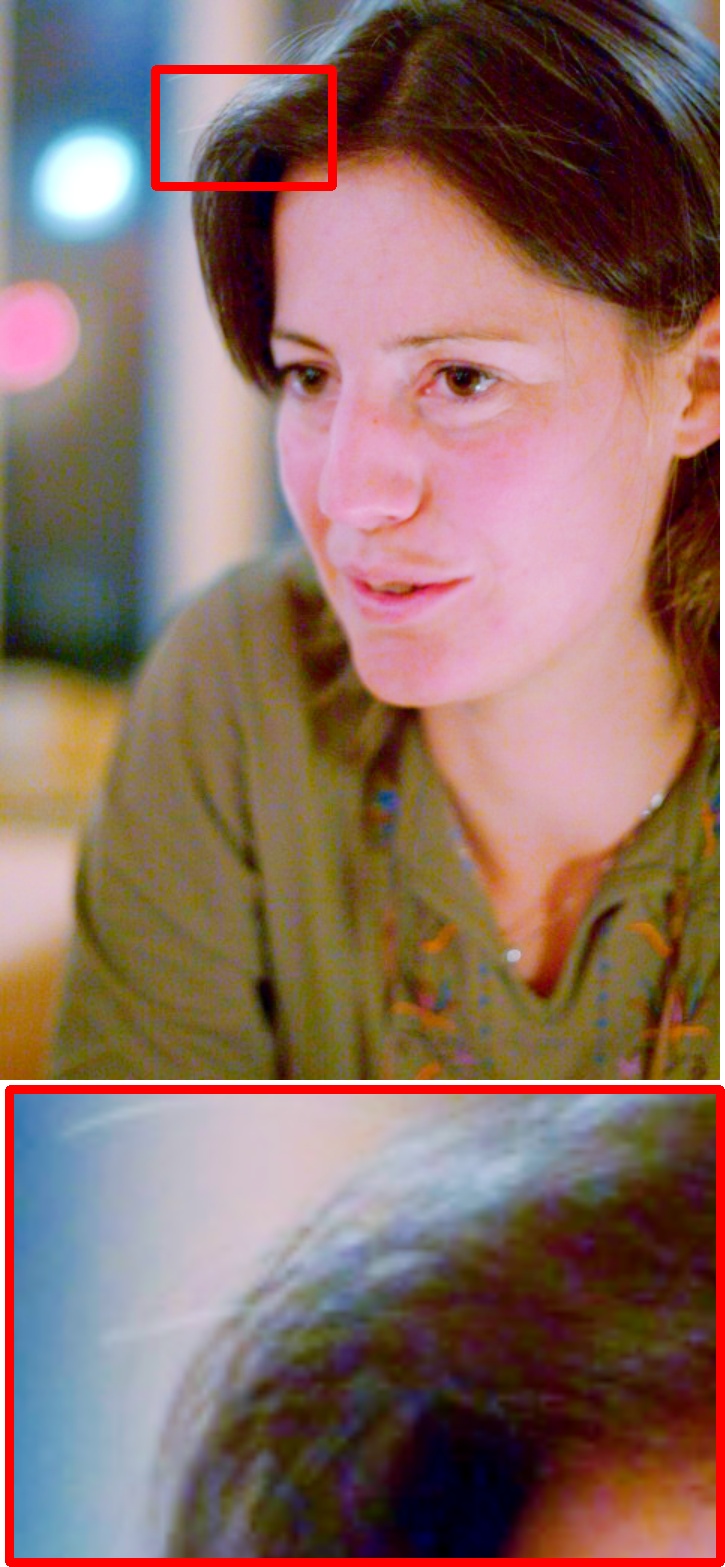}}
            \centerline{(d) CSRNet}\medskip
        \end{minipage}
        \begin{minipage}[b]{.12\linewidth}
            \centering
            \centerline{\includegraphics[height=4.35cm, frame]{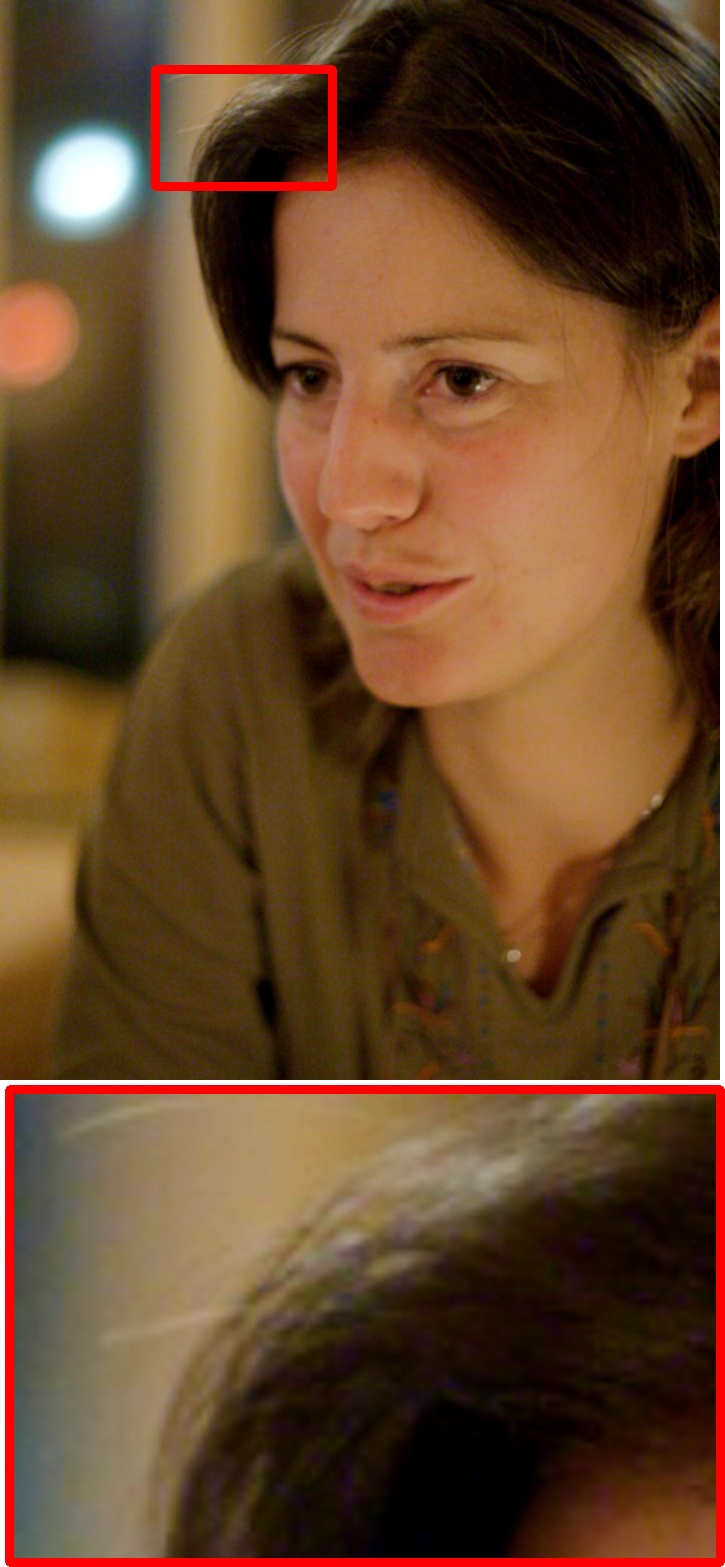}}
            \centerline{(e) 3D-LUT}\medskip
        \end{minipage}
        \begin{minipage}[b]{0.12\linewidth}
            \centering
            \centerline{\includegraphics[height=4.35cm, frame]{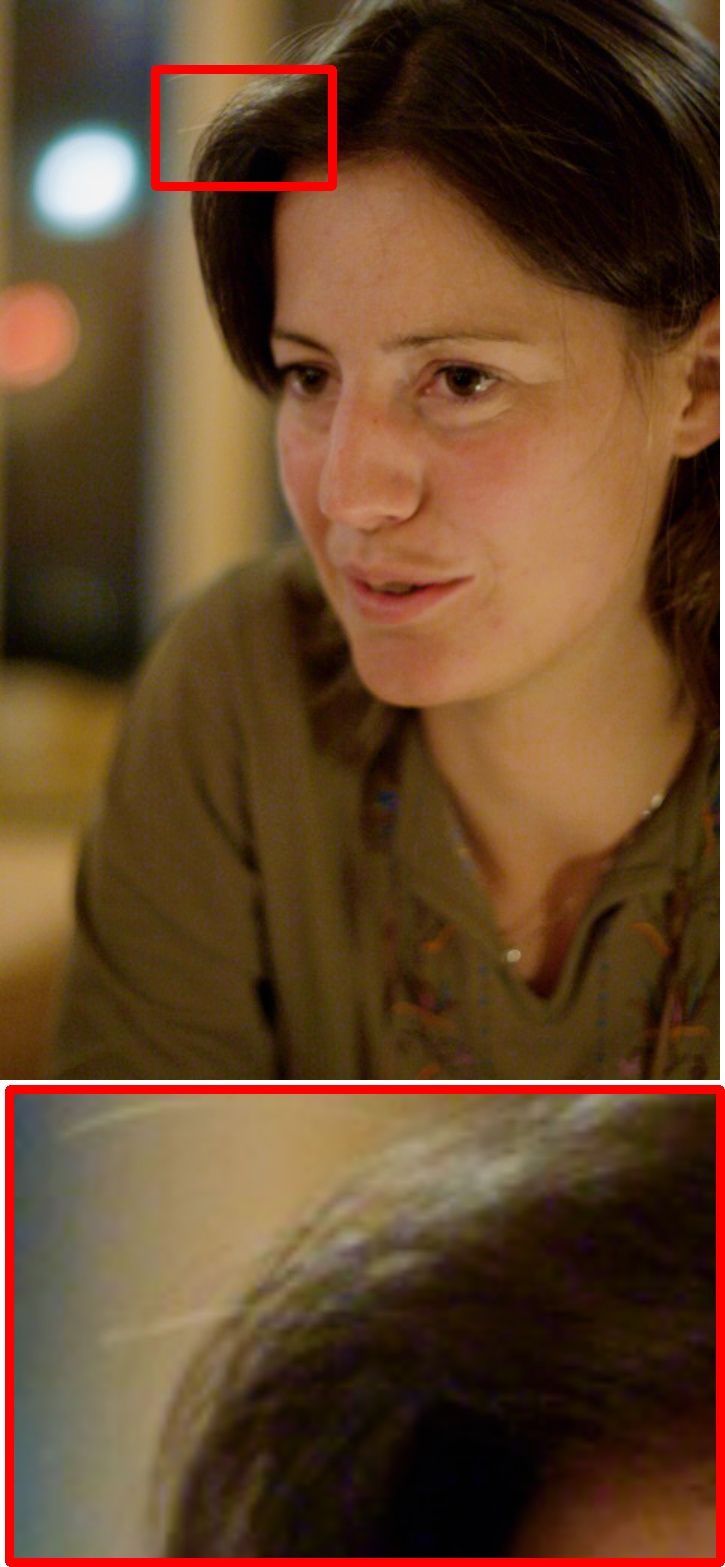}}
            \centerline{(f) SepLUT}\medskip
        \end{minipage}
        \begin{minipage}[b]{0.12\linewidth}
            \centering
            \centerline{\includegraphics[height=4.35cm, frame]{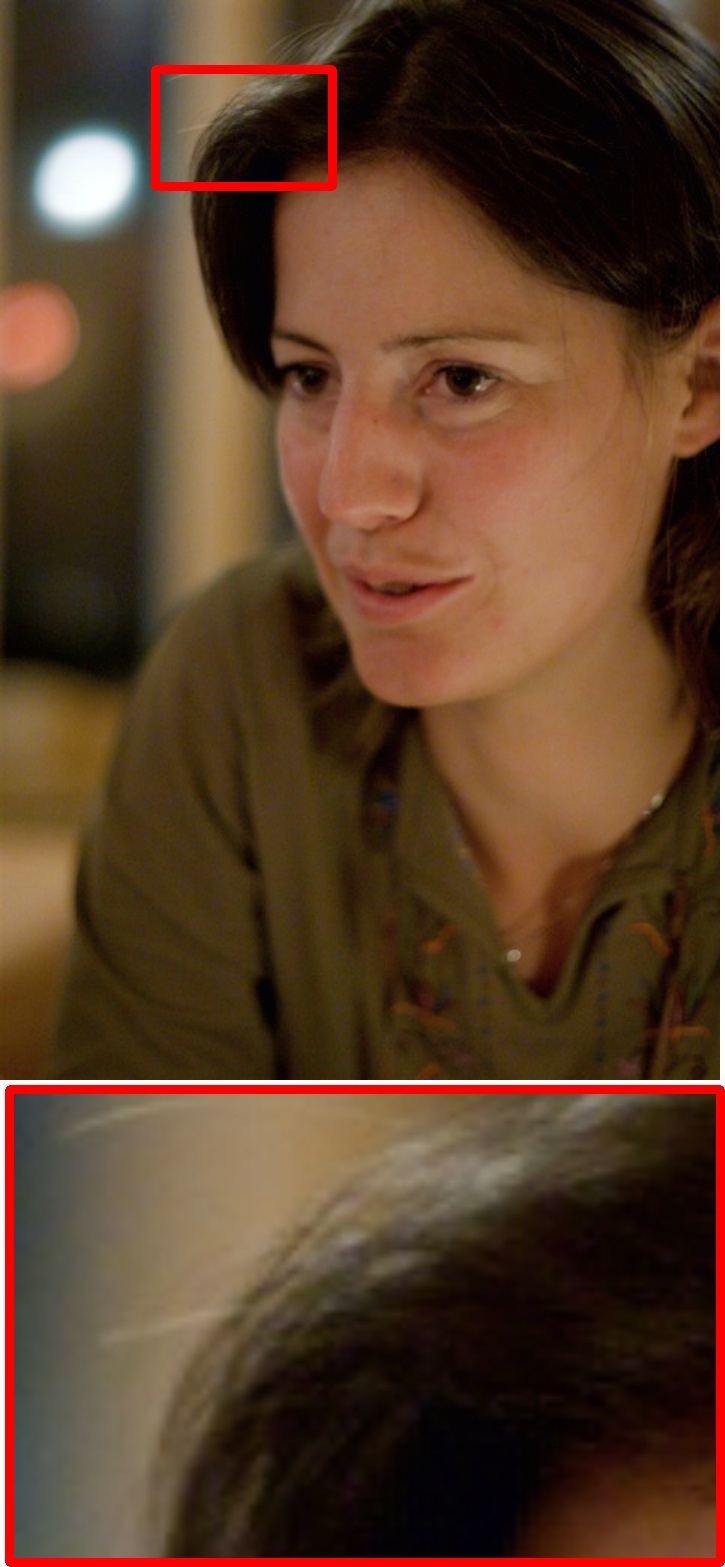}}
            \centerline{(g) Ours}\medskip
        \end{minipage}
        \begin{minipage}[b]{0.12\linewidth}
            \centering
            \centerline{\includegraphics[height=4.35cm, frame]{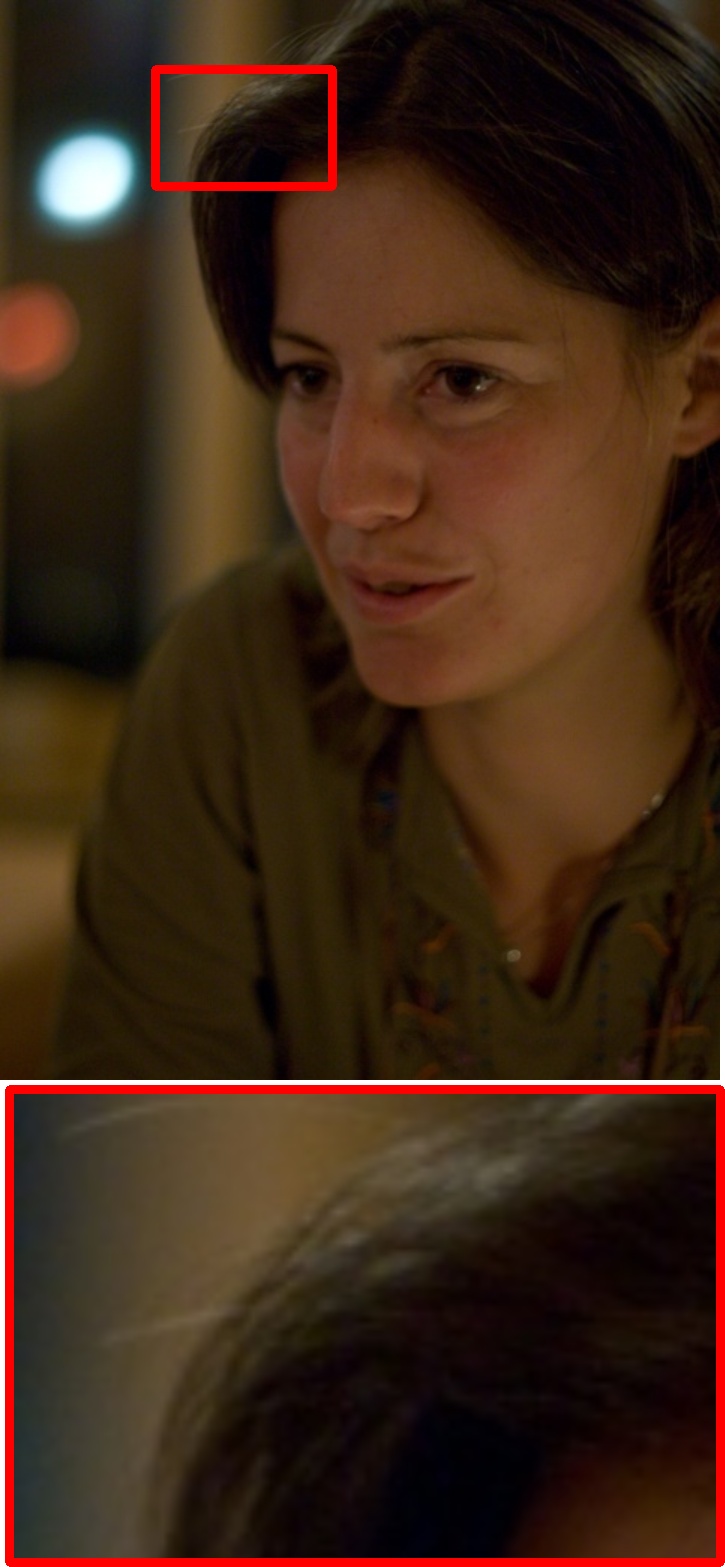}}
            \centerline{(h) Target}\medskip
        \end{minipage}
    \end{minipage}

    \begin{minipage}[b]{1.0\linewidth}
        \begin{minipage}[b]{0.24\linewidth}
            \centering
            \centerline{\includegraphics[height=2.65cm, frame]{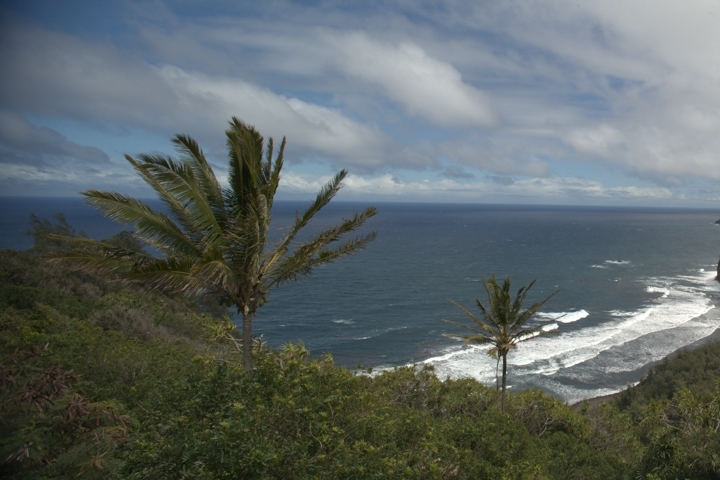}}
            \centerline{(a) Input}\medskip
        \end{minipage}
        \hfill
        \begin{minipage}[b]{0.24\linewidth}
            \centering
            \centerline{\includegraphics[height=2.65cm, frame]{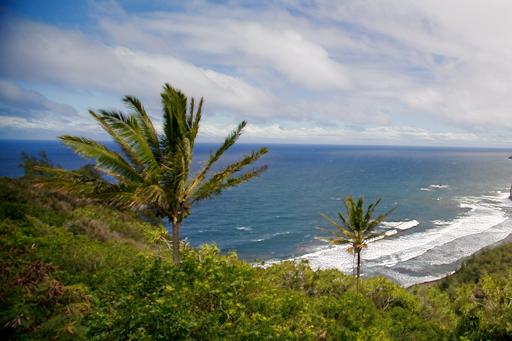}}
            \centerline{(b) DPE}\medskip
        \end{minipage}
        \hfill
        \begin{minipage}[b]{0.24\linewidth}
            \centering
            \centerline{\includegraphics[height=2.65cm, frame]{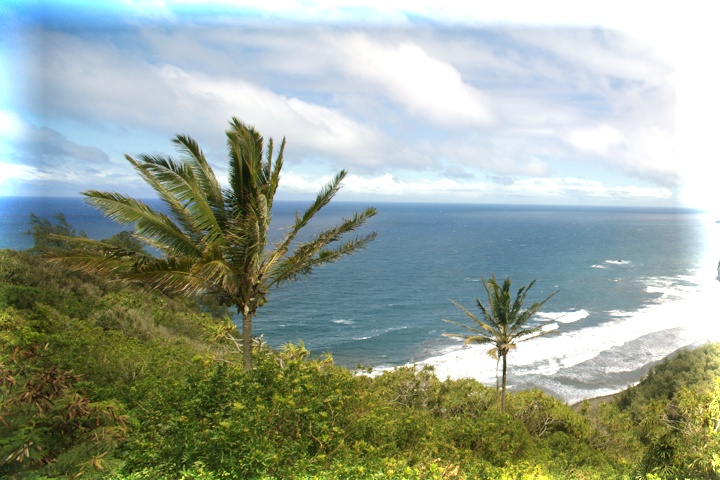}}
            \centerline{(c) UPE}\medskip
        \end{minipage}
        \hfill
        \begin{minipage}[b]{0.24\linewidth}
            \centering
            \centerline{\includegraphics[height=2.65cm, frame]{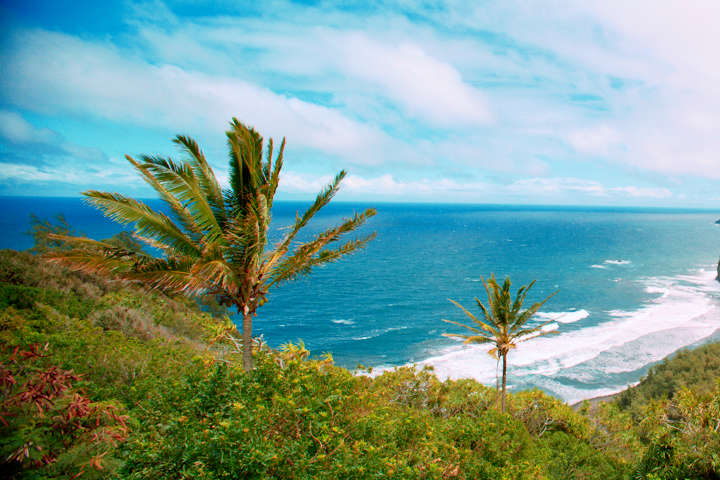}}
            \centerline{(d) CSRNet}\medskip
        \end{minipage}
    \end{minipage}   

    \begin{minipage}[b]{1.0\linewidth}
        \begin{minipage}[b]{0.24\linewidth}
            \centering
            \centerline{\includegraphics[height=2.65cm, frame]{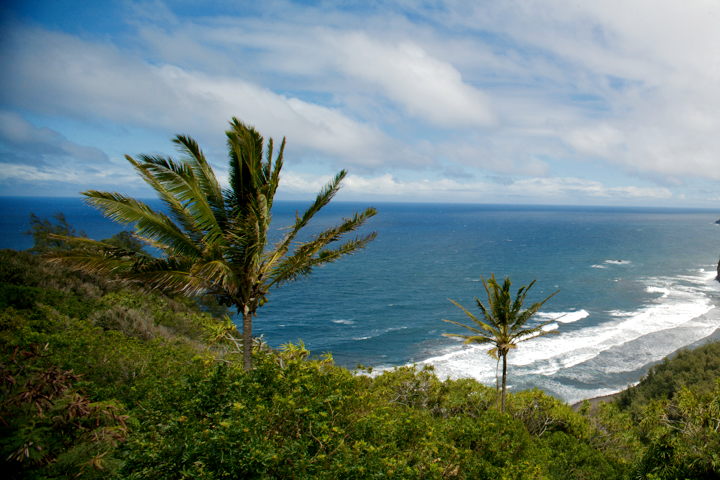}}
            \centerline{(e) 3D-LUT}\medskip
        \end{minipage}
        \hfill
        \begin{minipage}[b]{0.24\linewidth}
            \centering
            \centerline{\includegraphics[height=2.65cm, frame]{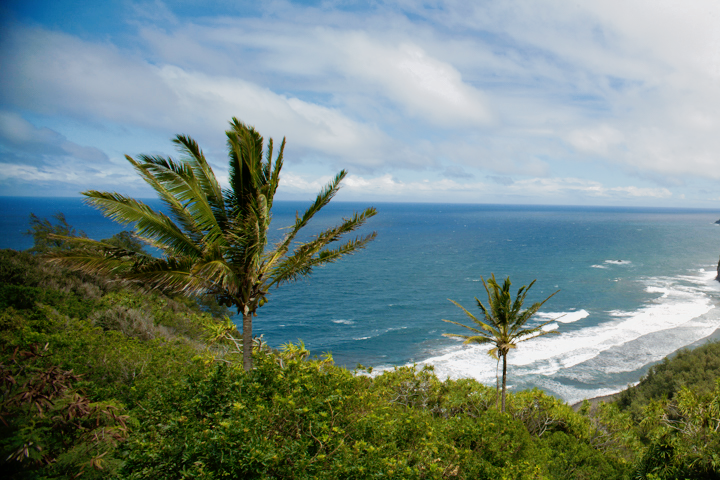}}
            \centerline{(f) SepLUT}\medskip
        \end{minipage}
        \hfill
        \begin{minipage}[b]{0.24\linewidth}
            \centering
            \centerline{\includegraphics[height=2.65cm, frame]{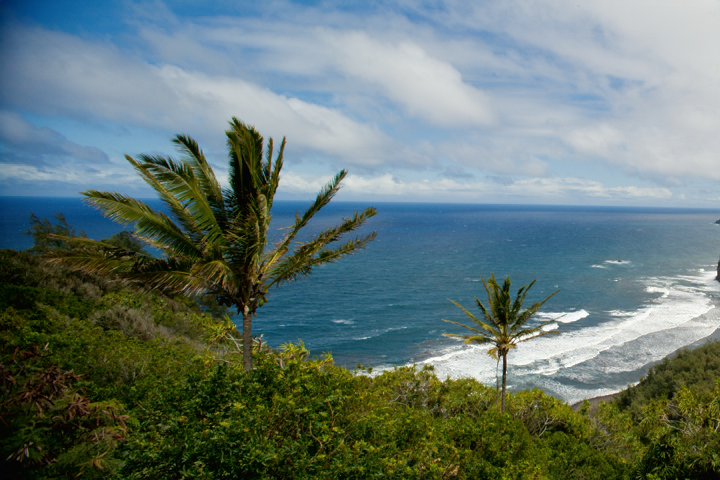}}
            \centerline{(g) Ours}\medskip
        \end{minipage}
        \hfill
        \begin{minipage}[b]{0.24\linewidth}
            \centering
            \centerline{\includegraphics[height=2.65cm, frame]{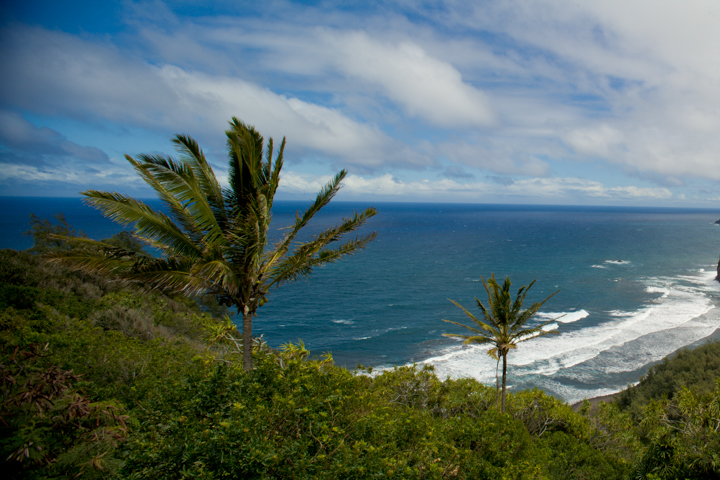}}
            \centerline{(h) Target}\medskip
        \end{minipage}
    \end{minipage}
    \caption{
    Visual comparison of different methods for image enhancement on the FiveK dataset. Our results are visually better in color tone and details. DPE, UPE and CSRNet results differ significantly from the target. The color, exposure and detail reproduction performances of them are not satisfying. 3D-LUT and SepLUT perform better, but the tone mapping is not good enough, which is often brighter or darker than the target.
    }
    \label{fig:results}
\end{figure*}

\section{EXPERIMENTS and results}

\subsection{Experimental setup}

We used two datasets for training and evaluation, MIT-Adobe FiveK~\cite{bychkovsky2011learning} and HDR+~\cite{hasinoff2016burst}. The MIT-Adobe FiveK dataset is currently the largest image enhancement dataset consisting of five retouched copies of 5,000 original photos from various contexts. The 3640-scene HDR+ dataset for high dynamic range and low-light imagery is a burst photography dataset from the Google camera group. For a fair comparison, we use the same dataset configuration as 3D-LUT~\cite{zeng2020learning} and transform all images to the more common 480p resolution and standard PNG format.

We evaluate different methods using three metrics, including Peak Signal-to-Noise Ratio~(PSNR), Structural Similarity~(SSIM) and $\Delta E$. $\Delta E$ is a measurement of human-perceived color variation in the CIELab color space~\cite{backhaus2011color}.
A greater value of PNSR and SSIM implies a better performance, whereas a lower value for $\Delta E$ indicates improved color appearance.

\subsection{Comparisons with state-of-the-art methods}
A total of eight state-of-the-art methods are compared with us: HDRNet~\cite{gharbi2017deep}, DPE~\cite{chen2018deep}, UPE~\cite{wang2019underexposed}, CSRNet~\cite{he2020conditional}, DeepLPF~\cite{moran2020deeplpf}, 3D-LUT~\cite{zeng2020learning}, STAR-DCE~\cite{zhang2021star} and SepLUT~\cite{yang2022seplut}. Table~\ref{table:exp_1} displays the quantitative results in terms of PSNR, SSIM, and $\Delta E$. As shown in the table, our method outperforms others considerably in terms of all metrics. Note that N/A indicates that the result is unavailable due to the absence of source code or report. When feasible, we re-evaluated each model using their available pretrained models on both datasets. 

Furthermore, qualitative results are displayed in Figure~\ref{fig:results}. Our model creates superior visual effects, and we not only preserve the color tone but also restore the finer details to their original state.

\subsection{Ablation studies}
In this section, we conduct ablation studies to investigate the effects and selection of our modules. As shown in Table~\ref{table:exp_1}, GSR stands for our global stylization remapping module and DPR stands for detailed parametric refinement module. Here we provide two modules as alternatives to GSR and DPR, ConvNet and UNet~\cite{ronneberger2015u}. ConvNet~\cite{cun2020towards} is a popular cascaded aggregation network that has shown effective for image restoration as a substitution to GSR, while UNet employs the same downsampling pattern as GSR.

The first half of the ablation experiments are used to improve the low-frequency component, while the second half are utilized for global optimization. Our UNet block for the high-frequency domain refinement remains unchanged.

Initial attempts to replace DPR with UNet for global optimization resulted in declining metrics. Next, we replaced GSR with ConvNet and coupled it with DPR and UNet, but the metrics were not as robust as they were with our model, and they decreased significantly with ConvNet and UNet. We also replaced GSR with UNet and paired it with DPR and UNet, but the results were not comparable to our model. In conclusion, the optimal combination is GSR and DPR, which delivers the highest metrics and the best visual impression.

\section{CONCLUSION}
In this paper, we present a transformer-based model in the wavelet domain for refining an image's various frequency bands. We thoroughly optimize the low frequency space of the image using DWT and transformer. The result of the transformer is IDWT with the UNet-optimized high frequency domain before being input into our Global stylization remapping module to further improvement. Our strategy concentrates on both local and global level for improvement, which is demonstrated to provide superior results. In the future, we will try to increase the downsampling multiplier of wavelet pooling, and introduce a more effective attention mechanism.

\bibliographystyle{IEEEbib}
\bibliography{ref}

\end{document}